\documentclass{article}
\usepackage{iclr2027_conference,times}

\usepackage{amsmath,amssymb,amsfonts}
\usepackage{algorithmic}
\usepackage{algorithm}
\usepackage{float}
\floatstyle{ruled}
\restylefloat{algorithm}
\usepackage{graphicx}
\usepackage{textcomp}
\usepackage{xcolor}
\usepackage{CJKutf8}
\usepackage{multirow}
\usepackage{siunitx}
\usepackage{booktabs}
\usepackage{colortbl}
\usepackage{wrapfig}
\usepackage{pdflscape}
\usepackage{hyperref}
\usepackage{url}
\usepackage{svg}
\usepackage{enumitem}
\usepackage{array}
\usepackage{pifont}

\definecolor{bestpurple}{RGB}{230,210,250}
\definecolor{secondblue}{RGB}{210,230,250}
\definecolor{headergray}{RGB}{235,235,235}
\definecolor{countred}{RGB}{255,220,220}

\newcommand{\best}[1]{%
\cellcolor{bestpurple}\textbf{#1}
}

\newcommand{\second}[1]{%
\cellcolor{secondblue}\underline{\textbf{#1}}
}

\newcommand{\rowcg}{%
\rowcolor{headergray}
}
\def\BibTeX{{\rm B\kern-.05em{\sc i\kern-.025em b}\kern-.08em
    T\kern-.1667em\lower.7ex\hbox{E}\kern-.125emX}}

\title{QiYao-I: A Manifold Based Foundation Model for Irregular Multivariate Time Series Forecasting}

\author{
\textbf{Linfeng Wang}$^{1}$
\textbf{Ruitong Zhang}$^{1}$
\textbf{Kai Zhao}$^{1}$
\textbf{Yang Shu}$^{1}$
\textbf{Zhongwen Rao}$^{2}$ 
\textbf{Meng Wang}$^{2}$ \\
\textbf{Yijie Li}$^{2}$
\textbf{Bin Yang}$^{1}$
\textbf{Chenjuan Guo}$^{1,\ast}$ \\
\\[-1mm]
$^{1}$East China Normal University
$^{2}$Huawei Technologies Ltd. \\
$^{1}$\{lfwang, rtzhang\}@stu.ecnu.edu.cn \\
$^{1}$\{kzhao, yshu, byang, cjguo\}@dase.ecnu.edu.cn \\
$^{2}$\{raozhongwen, wangmeng71, liyijie5\}@huawei.com \\
$^{\ast}$Corresponding author
}

\iclrfinalcopy 

\begin{document}

\maketitle

\begin{abstract}
Irregular multivariate time series forecasting is a challenging yet important problem in real-world applications, where observations are often irregularly sampled and asynchronously recorded across variables. Existing time series foundation models are mostly built on regularly sampled sequences, making them difficult to generalize to irregular time intervals and asynchronous cross-variable dependencies. To address these challenges, we propose \textbf{QiYao-I}, a manifold based foundation model for irregular multivariate time series forecasting. Specifically, we
introduce a novel sampling-conditioned temporal manifold attention mechanism that maps real timestamps into a learnable temporal manifold feature space and injects temporal manifold biases into attention layers, enabling the model to capture both irregular time intervals and local sampling structures. Further, we propose a dynamic
variable interaction mechanism with frequency awareness. It selectively performs cross-variable message passing under asynchronous observations. Extensive experiments on real-world irregular multivariate forecasting benchmarks demonstrate that QiYao-I achieves superior performance compared with both time series foundation models and end-to-end irregular forecasting models, showing strong generalization ability in zero-shot and few-shot settings. 

\end{abstract}


\section{Introduction}

In recent years, foundation models have been reshaping the paradigm of time series modeling. Unlike traditional models designed for a single dataset or a specific task, time series foundation models are typically pretrained on large-scale data to learn transferable and generalizable temporal representations, enabling unified modeling across diverse forecasting tasks. 

\begin{wrapfigure}{R}{0.62\textwidth}
\vspace{-4mm}
\centering
\includegraphics[width=\linewidth]{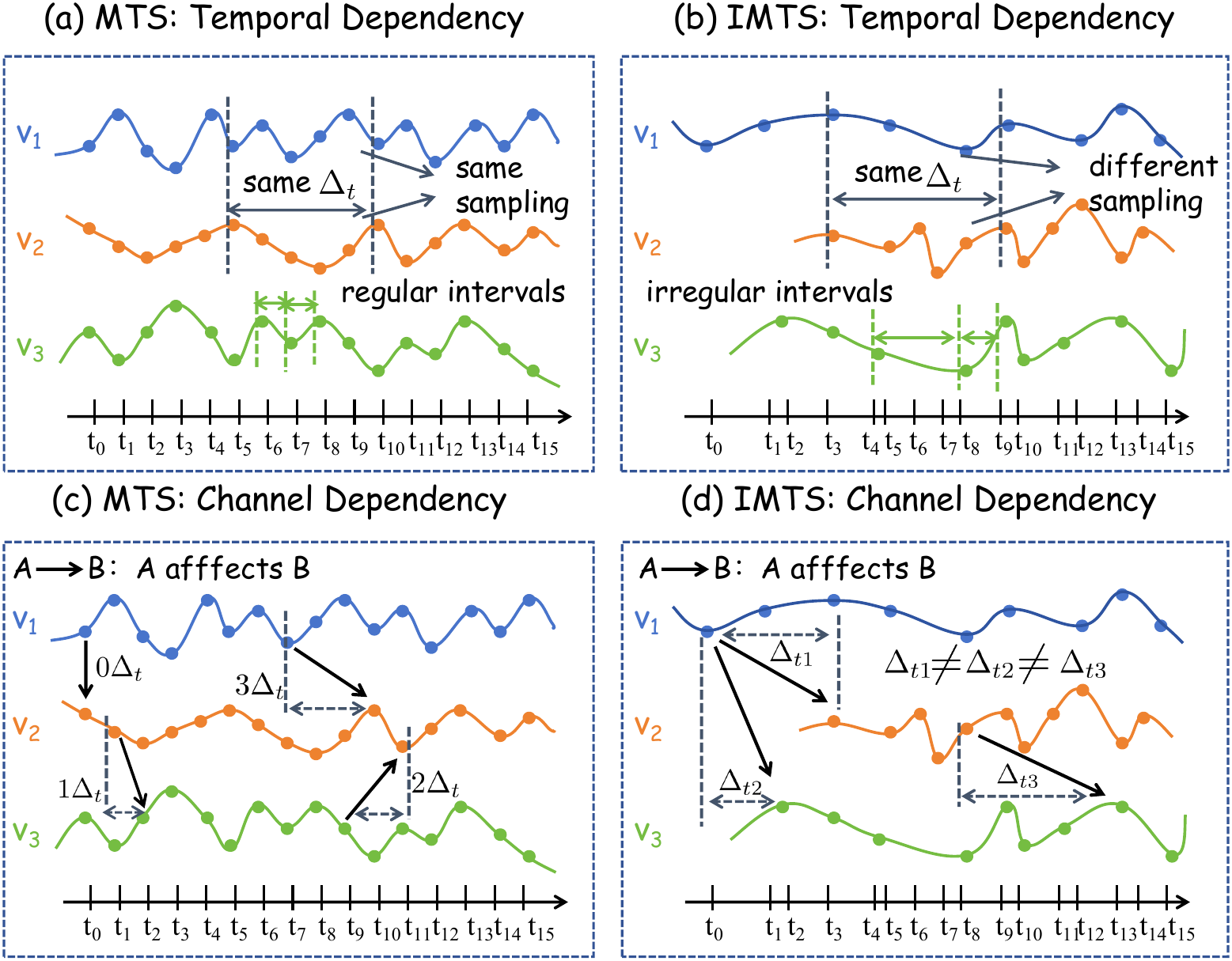}
\caption{The difference between regular multivariate time series (MTS) and irregular multivariate time series (IMTS).}
\label{intro1}
\vspace{-5mm}
\end{wrapfigure}

However, most existing time series foundation models assume regular multivariate time series (MTS), where two pairs of observations are equally spaced and same time intervals $\Delta_t$ contain the same number of sampled points, as shown in Fig.~\ref{intro1}(a). Similarly, different variables are observed across fixed regular temporal intervals, enabling cross-variable interactions either at the
same timestamp or across multiple sampling intervals whose durations are the same, as shown in Fig.~\ref{intro1}(c)).

Nevertheless, this assumption does not hold in many real world scenarios. Real-world systems often produce irregular multivariate time series (IMTS). For example, sensor failures may lead to irregular intervals, and different clinical procedures yield variable-specific asynchronous observations. As shown in Fig.~\ref{intro1}(b), adjacent observations in IMTS have irregular
intervals. Even if two pairs of observations span the same time gap $\Delta_t$, their intervals have different numbers of sampled points. Moreover, Fig.~\ref{intro1}(d) shows that different variables are observed at variable-specific irregular timestamps, so that cross-variable effects occur distinctively across time intervals.

Although existing time series foundation models have shown strong generalization ability in regularly sampled settings, they still rely on regular temporal structures and aligned variable observations, making direct transfer to IMTS challenging.

\textbf{Challenge 1: Limited generalization to irregular time steps.}
Most mainstream time series foundation models~\citep{khwaja2026toto,ansari2025chronos,liu2025sundial,das2023decoder,liu2024timer} treat observations as equally spaced tokens and use discrete positional embeddings. These embeddings capture token order but not actual time intervals: under irregular sampling, two observation pairs with the same token order may correspond to very different real time intervals, leading to incorrect dependency learning and biased generalization. Although MIRA~\citep{li2026mira} introduces scalar timestamps into positional embeddings, it does not fully account for local sampling structure. As illustrated in Fig.~\ref{intro1}(b), the same time interval can imply different dependencies in densely and sparsely sampled regions. Therefore, simply introducing scalar timestamps is not enough to fully address the generalization problem under irregular time steps.

\textbf{Challenge 2: Limited cross-variable interaction under asynchronous observations.}
Channel-independent models~\citep{liu2025sundial,das2023decoder,liu2024timer,wang2025lightgts,wang2025rose} process variables separately, failing to capture cross-variable dependencies. Channel-dependent models~\citep{khwaja2026toto,ansari2025chronos,woo2024unified} enable such exchange but typically assume observations are aligned within shared time intervals, as shown in Fig.~\ref{intro1}(c). In IMTS, variables are observed asynchronously, and dependencies can span distinct intervals with different sampling structures, as shown in Fig.~\ref{intro1}(d). As a result, existing time series foundation models still lack a transferable interaction mechanism that can capture asynchronous cross-variable dependencies.

\textbf{To address the challenge of irregular temporal modeling in foundation model generalization}, we introduce \textit{a novel sampling-conditioned temporal manifold attention mechanism}. Specifically, the temporal manifold refers to a learnable space that embeds scalar timestamps. Instead of relying on dataset-specific interpolation, this mechanism maps scalar timestamps into a learnable temporal manifold. It models temporal relations by learning manifold
distances and converts these distances into temporal manifold biases. Furthermore, the model introduces a Gaussian-kernel descriptor to distinguish observation pairs, which span the same time interval but have different numbers of sampled points by adaptively adjusting the bandwidth of these temporal manifold biases. By injecting temporal manifold biases into attention logits, the model can learn transferable temporal dependencies across heterogeneous IMTS data with different sampling structures.
\textbf{To address the challenge of asynchronous cross-variable modeling for foundation models}, we propose \textit{a frequency-guided dynamic variable interaction mechanism}. Observations from different variables are not aligned, and their dependencies involve multiple points within a local interval. We therefore group observations into patches for cross-variable interaction. To identify related patch intervals across
asynchronously observed variables, we compute temporal distances between patches and select the top-$K$ nearest patches from other variables. However, being close in time does not indicate whether their observed values follow similar patterns, so giving all selected patches equal weight may introduce unrelated information. We therefore use learnable spectrum projection and aggregation to extract patch-level and global-level frequency features directly from irregular values and timestamps. Patch-level features capture local patterns within each selected patch, while variable-level features summarize broader patterns. Their similarities jointly determine the aggregation weights to emphasize information from patches with similar local and global temporal patterns.

To this end, we propose \textbf{QiYao-I}, a unified foundation model framework designed for irregular multivariate time series. Our main contributions are summarized as follows:
\begin{itemize}[leftmargin=20pt]
\vspace{-3mm}
    \item We propose QiYao-I, which jointly models irregular temporal dependencies and asynchronous cross-variable interactions in a unified framework.

  \item We propose \textit{a novel sampling-conditioned temporal manifold time attention}, which captures irregular time intervals and distinguishes the same interval under different sampling structure.
  
    \item We propose \textit{a frequency-guided dynamic variable interaction mechanism} that selects patches from other variables based on temporal distance and weights their contributions by frequency similarity, enabling selective interactions across asynchronous variables.

    \item Extensive experiments show that QiYao-I achieves superior forecasting performance and strong generalization ability in both zero-shot and few-shot settings.
\end{itemize}

\section{Related Works}

\textbf{End-to-End Irregular Time Series Modeling.}
Irregular time series modeling aims to handle observations sampled at uneven and asynchronous timestamps. Existing end-to-end methods mainly include time-aware recurrent models~\citep{DBLP:journals/corr/ChePCSL16,baytas2017patient}, continuous-time models Latent ODE~\citep{rubanova2019latent}, Neural CDE~\citep{kidger2020neural}, as well as set-, graph-, attention-, and patch-based models~\citep{DBLP:conf/icml/HornMBRB20,zhang2021graph,DBLP:conf/iclr/ShuklaM21,DBLP:conf/icml/ZhangYL0024}. These methods improve the modeling of irregular observations from different perspectives and often perform well when trained on a specific target dataset. However, most of them are task-specific models and are not designed for large-scale pretraining or zero-shot transfer. Different from these methods, QiYao-I is built as a foundation model for irregular multivariate time series. It directly models irregular timestamps in a learnable temporal manifold.

\textbf{Time Series Foundation Models.}
Recent time series foundation models have achieved promising zero-shot and few-shot forecasting performance through large-scale pretraining. Representative models include Toto2~\citep{khwaja2026toto}, Chronos2~\citep{ansari2025chronos},
TimesFM2.5~\citep{das2023decoder}, MOIRAI~\citep{woo2024unified}, and MoE-based models such as Time-MoE and MOIRAI-MoE~\citep{shi2025time,liu2024moirai}. However, most existing foundation models are designed for regularly sampled time series. They usually regard a sequence as equally spaced tokens and use discrete positional encodings to model temporal order. When applied to irregular time series, these models often require resampling or interpolation, which may change the original sampling structure and introduce extra bias. Although MIRA~\citep{li2026mira} introduces continuous-time modeling for irregular medical time series, how to model local sampling structure remains underexplored. In contrast, QiYao-I supports irregular inputs from pretraining to inference. By injecting manifold-distance biases into attention, QiYao-I can better capture irregular time intervals and transfer across datasets with different sampling structure.


\section{QiYao-I}
\label{sec:method}

\subsection{Framework Overview}
We develop \textbf{QiYao-I}, a foundation model for irregular multivariate time series forecasting. An irregular multivariate time series (IMTS) contains $V$ variables observed at uneven and asynchronous timestamps. After padding, its historical observations are represented by values $\mathbf{X}^h\in\mathbb{R}^{V\times L_h}$, timestamps $\mathbf{T}^h\in\mathbb{R}^{V\times L_h}$, and validity masks $\mathbf{M}^h\in\{0,1\}^{V\times L_h}$. Forecasting queries are specified by future timestamps $\mathbf{T}^p\in\mathbb{R}^{V\times L_p}$ and masks $\mathbf{M}^p\in\{0,1\}^{V\times L_p}$. Here, $L_h$ and $L_p$ are the maximum historical and future lengths; the masks indicate valid observations and prediction queries, respectively. The forecasting objective is to learn
$
\widehat{\mathbf{X}}^p
=
f_\theta\left(
\mathbf{X}^h,\mathbf{T}^h,\mathbf{M}^h,
\mathbf{T}^p,\mathbf{M}^p
\right),
\label{eq:forecast_problem}
$

\begin{figure}[t]
\centering
\includegraphics[width=\linewidth]{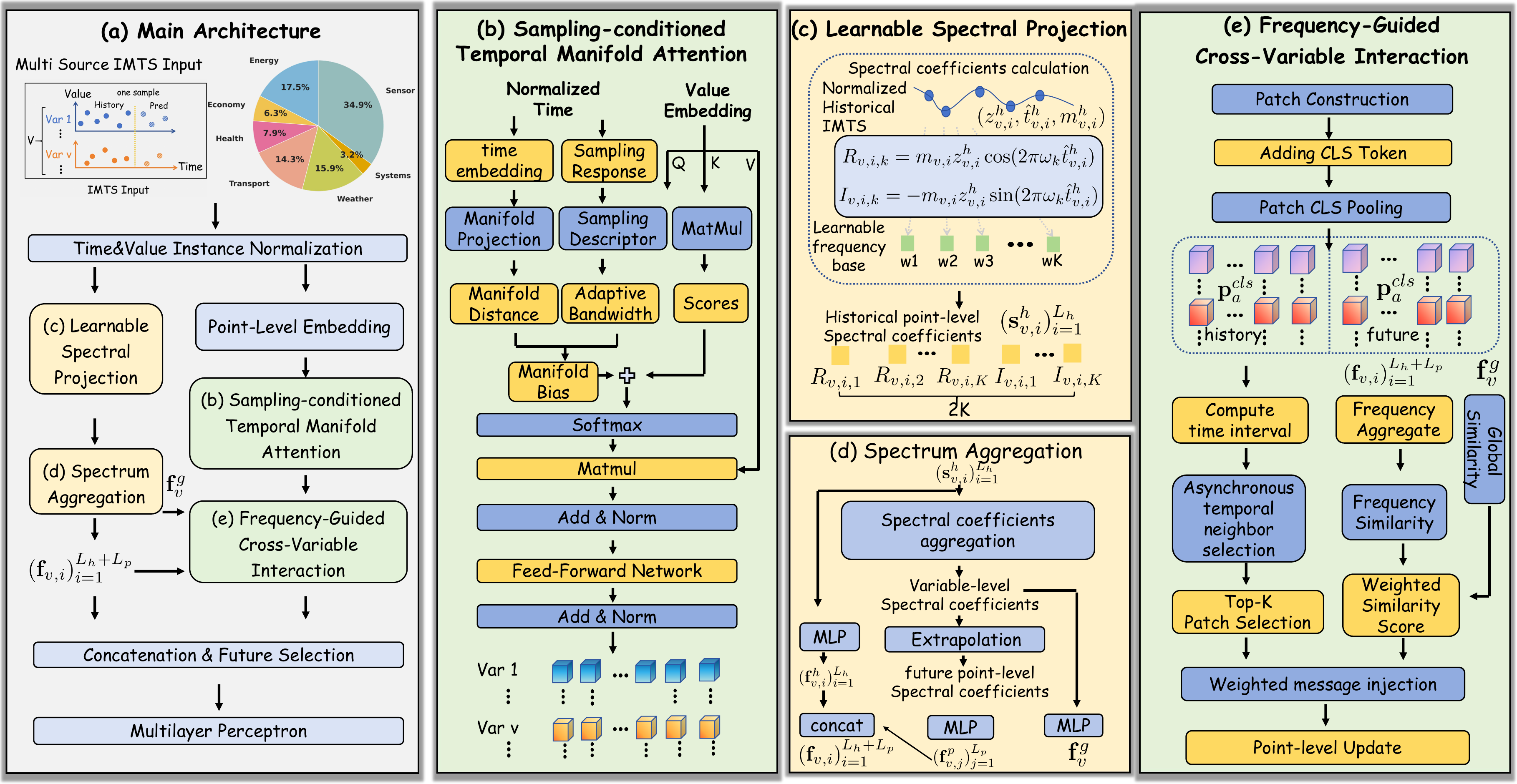}
\caption{Overall architecture of QiYao-I.}
\label{fig:QiYao-I}
\vspace{-5mm}
\end{figure}

As illustrated in Figure~\ref{fig:QiYao-I}, three components model irregular temporal dependencies and asynchronous cross-variable interactions for cross-dataset generalization. First, \textit{Sampling-Conditioned Temporal Manifold Attention} maps timestamps into a learnable temporal manifold and constructs temporal manifold biases from pairwise manifold distances to model irregular temporal dependencies under different sampling structures. Second, to capture periodic patterns and guide selective patch-level
interaction, \textit{Learnable Irregular Spectral Projection and Spectrum Aggregation} extracts frequency representations directly from irregular historical observations and extrapolates these representations to future query timestamps. Third, \textit{Frequency-Guided Cross-Variable Interaction} selects temporally relevant patches from other variables and weights them according to local and global frequency similarity to model asynchronous cross-variable interaction. Finally, each future temporal state is fused with its extrapolated frequency representations to produce a forecast.

\subsection{Sampling-Conditioned Temporal Manifold Attention}
\label{sec:continuous_time_encoder}

To model irregular temporal dependencies, prior work uses continuous-time RoPE~\citep{su2024roformer,li2026mira} to model irregular dependency, but its rotary transformations do not describe the local sampling structure. We introduce sampling-conditioned temporal manifold attention, which maps timestamps into a shared learnable space, converts pairwise distances into attention biases, and adapts their
bandwidth to each input's local sampling structure. Through multi-source pre-training, the model learns generalizable patterns across different sampling structures, allowing temporal biases to adapt to diverse observation patterns without dataset-specific temporal interpolation.



\paragraph{Bandwidth concept.}
For a Gaussian kernel
$K_{\sigma}(\delta)=\exp[-\delta^2/(2\sigma^2)]$,
its logarithm gives a negative distance penalty.
A larger bandwidth $\sigma$ weakens this penalty at a fixed
distance. 
Our  temporal bias follows this log-kernel form,
with a bandwidth conditioned on local sampling structure.

\paragraph{Normalization and token embedding.}
For each variable $v\in[V]$, we compute the historical mean
$\mu_v$ and standard deviation $s_v$ from valid observations only, and normalize values as $z_{v,i}=(x^h_{v,i}-\mu_v)/s_v$.
We use $s_v=1$ for insufficient  history and
$\mu_v=0$ when history is empty. Following the future masked token formulation~\citep{ansari2025chronos}, we represent each future position with a learnable mask token. The historical and future tokens are initialized as:
\begin{equation}
\mathbf{c}^{h,(0)}_{v,i}
=
\mathbf{W}_x\operatorname{asinh}(z_{v,i})+\mathbf{b}_x,
\qquad
\mathbf{c}^{p,(0)}_{v,j}
=
\mathbf{e}_{\mathrm{mask}},
\label{eq:token_embedding}
\end{equation}
where $i\in[L_h]$, $j\in[L_p]$, $\mathbf{W}_x\in\mathbb{R}^{d\times1}$, and $\mathbf{b}_x,\mathbf{e}_{\mathrm{mask}}\in\mathbb{R}^{d}$.

To place timestamps from heterogeneous datasets on a common scale, each historical or future timestamp is normalized as $\widehat{t}=R(t-t_0)/(t_b-t_0)$, where $t_0$ and $t_b$ denote the start and historical boundary of the sampled window, respectively, and $R>0$ is a fixed reference scale. This transformation maps the historical window to $[0,R]$, while allowing future timestamps to exceed $R$. It preserves relative temporal gaps and is unaffected by missing observations or the forecast horizon.

\paragraph{Learnable temporal manifold distance.}
Although continuous-time RoPE~\citep{li2026mira} injects scalar $\widehat{t}$ into attention queries and keys, this can not reflect the local sampling structures. We first map each normalized timestamp into a learnable temporal manifold. For layer $\ell\in[N_T]$ and component $k\in\{0,\ldots,d_t-1\}$, the timestamp embedding is:
\begin{equation}
    \mathbf{e}_{\ell}(\widehat{t}_u)[k]
    =
    \begin{cases}
        \omega_{\ell,0}\widehat{t}_u+\alpha_{\ell,0},
        & k=0,\\
        \sin(\omega_{\ell,k}\widehat{t}_u+\alpha_{\ell,k}),
        & 1\leq k<d_t,
    \end{cases}
    \qquad
    \mathbf{e}_{\ell}(\widehat{t}_u)\in\mathbb{R}^{d_t},
    \label{eq:time_embedding}
\end{equation}
where $N_T$ is the number of temporal layers, $d_t$ is the time-embedding dimension, and all $\omega_{\ell,k}$ and $\alpha_{\ell,k}$ are learnable.
For attention head $r\in[H]$, an MLP $\phi_{\ell,r}:\mathbb{R}^{d_t}\rightarrow\mathbb{R}^{d_m}$ produces a head-specific manifold representation. For positions $u,w\in[L_h+L_p]$, their manifold distance is:
\begin{equation}
    D^{\ell,r}_{uw}
    =
    \left\|
        \phi_{\ell,r}(\mathbf{e}_{\ell}(\widehat{t}_u))
        -
        \phi_{\ell,r}(\mathbf{e}_{\ell}(\widehat{t}_w))
    \right\|_2^2,
    \label{eq:manifold_distance}
\end{equation}
Here, $H$ is the number of attention heads and $d_m$ is the
representation dimension. Temporal manifold denotes the learned timestamp representation space; $D^{\ell,r}_{uw}$
is its squared Euclidean distance.

\paragraph{Sampling-conditioned bandwidth.}
The manifold distance $D^{\ell,r}_{uw}$ depends on the timestamp pair, and we condition the bandwidth of the distance-based attention bias on local sampling structure, so that the same distance can receive different penalties under
different observation patterns across datasets.

For position $u$ of variable $v$, let
$\bar{t}_u=\min(\widehat{t}_u,R)$, where $R$ is the normalized
historical boundary. A historical position examines observations
around its own time; a future position examines observations near
$R$. Within a preceding window of width
$W>0$, the valid historical observations are:
\begin{equation}
\mathcal{I}_u
=
\left\{
i\in[L_h]:
m^h_{v,i}=1,\;
0\leq\bar{t}_u-\widehat{t}^h_{v,i}\leq W
\right\}.
\label{eq:local_history_set}
\end{equation}
We place $A$ anchors $\{c_a\}_{a=1}^{A}$ uniformly over $[-1,0]$.
Each anchor measures the concentration of observations in a different
part of this window:
\begin{equation}
S_{u,a}
=
\log\!\left[
1+
\sum_{i\in\mathcal{I}_u}
\exp\!\left(
-\frac{1}{2\kappa^2}
\left(
\frac{\widehat{t}^h_{v,i}-\bar{t}_u}{W}-c_a
\right)^2
\right)
\right],
\quad a\in[A],
\label{eq:sampling_measure}
\end{equation}
where $\kappa>0$ controls the width of the Gaussian kernels. \textit{The sampling responses $S_{u,1},\ldots,S_{u,A}$} can therefore distinguish
observation pairs that span the same time interval but have different
local sampling structures.

Let $\widehat{t}^{\,\mathrm{last}}_u$ be the latest valid historical
time not exceeding $\bar{t}_u$, and let $\mathcal{I}^{\leq}_u$
contain all such valid historical observations. We form the \textit{sampling descriptor}:
\begin{equation}
\mathbf{s}_u
=
\left[
S_{u,1},\ldots,S_{u,A};
\log\!\left(1+\frac{(\widehat{t}_u-R)_+}{W}\right);
\log\!\left(
1+\frac{(\widehat{t}_u-\widehat{t}^{\,\mathrm{last}}_u)_+}{W}
\right);
\mathbb{I}(|\mathcal{I}^{\leq}_u|>0)
\right].
\label{eq:sampling_descriptor}
\end{equation}
The descriptor $\mathbf{s}_u\in\mathbb{R}^{A+3}$ combines local sampling responses with the query's distance beyond $R$, its elapsed time since the latest observation, and a history-availability indicator. If no preceding observation exists, the elapsed-time feature and availability indicator are both set to zero.

A layer- and head-specific MLP converts this descriptor into a
positive scale:
\begin{equation}
r^{\ell,r}_u
=
\exp\!\left[
\lambda\tanh\!\left(g_{\ell,r}(\mathbf{s}_u)\right)
\right],
\label{eq:adaptive_bandwidth}
\end{equation}
where $\lambda>0$ bounds its adaptation range. The scale adjusts
the penalty derived from the manifold distance:
\begin{equation}
b^{\ell,r}_{uw}
=
-\frac{\eta_{\ell,r}D^{\ell,r}_{uw}}
{\beta_{\ell,r}^{\,2}r^{\ell,r}_u r^{\ell,r}_w},
\qquad
\eta_{\ell,r}=\operatorname{softplus}(\gamma_{\ell,r}),
\quad
\beta_{\ell,r}=\operatorname{softplus}(\sigma_{\ell,r})+\epsilon.
\label{eq:manifold_bias}
\end{equation}
The effective bandwidth is
$\beta_{\ell,r}\sqrt{r^{\ell,r}_u r^{\ell,r}_w}$. Consequently, two
pairs with the same manifold distance can receive different attention
biases when their local sampling structures differ.
\paragraph{Temporal manifold attention.}
Let $\mathbf{q}^{\ell,r}_u,\mathbf{k}^{\ell,r}_u\in\mathbb{R}^{d_r}$ be the query and key of head $r$, where $d_r=d/H$. Continuous-time RoPE and the manifold bias are combined as:
\begin{align}
    \widetilde{\mathbf{q}}^{\,\ell,r}_u
    &=
    \mathcal{R}(\widehat{t}_u)\mathbf{q}^{\ell,r}_u,
    &
    \widetilde{\mathbf{k}}^{\,\ell,r}_w
    &=
    \mathcal{R}(\widehat{t}_w)\mathbf{k}^{\ell,r}_w,
    \nonumber\\[-1mm]
    a^{\ell,r}_{uw}
    &=
    \frac{
        (\widetilde{\mathbf{q}}^{\,\ell,r}_u)^\top
        \widetilde{\mathbf{k}}^{\,\ell,r}_w
    }{\sqrt{d_r}}
    +
    b^{\ell,r}_{uw}
    +
    \mathcal{M}_{uw},
    &
    h^{\ell,r}_{uw}
    &=
    \operatorname{softmax}_{w}(a^{\ell,r}_{uw}),
    \label{eq:time_attention}
\end{align}
Here, $\mathcal{M}_{uw}=0$ for valid keys and $-\infty$ for
padding. Attention operates separately within each variable over its historical tokens and future query tokens, supporting generalization to datasets with different sampling structures. 

\subsection{Learnable Irregular Spectral Representation}
\label{sec:nudft_module}

Temporal attention models dependencies between observations within each variable. 
To capture periodic patterns that transfer across datasets and guide cross-variable interaction,
we learn a shared spectral mapping directly from irregular values and timestamps through multi-source pre-training. It provides local and global frequency
representations for comparing variables across different inputs.
Standard Discrete Fourier Transforms (DFT)~\citep{sundararajan2001discrete} assumes uniformly spaced observations. We instead evaluate sinusoidal bases at the actual normalized timestamps and learn their frequencies, obtaining spectral representations without resampling the input.

\paragraph{Learnable spectral projection.}
Standard DFT use sine and cosine bases on
a fixed frequency grid determined by the sampling interval and
sequence length. To extract frequency information from irregular observations, we
apply sine and cosine bases with learnable frequencies directly to their timestamps.
Let $K_f$ be the number of spectral components and
$w_k=\exp(\xi_k)>0$, $k\in[K_f]$, be learnable frequencies. These frequencies are defined on the normalized window-time scale. For variable $v\in[V]$ and historical position $i\in[L_h]$,
we compute the real and imaginary coefficients as:
\begin{equation}
[R^{\mathrm{re}}_{v,i,k},I^{\mathrm{im}}_{v,i,k}]
=
m^h_{v,i}z_{v,i}
[\cos(2\pi w_k\widehat{t}^h_{v,i}),
-\sin(2\pi w_k\widehat{t}^h_{v,i})],
\qquad k\in[K_f].
\label{eq:nudft}
\end{equation}
Following the real--imaginary decomposition of Fourier coefficients,
we concatenate both components into a real-valued spectral coefficients
$\mathbf{s}^h_{v,i}
=[\mathbf{R}^{\mathrm{re}}_{v,i};
\mathbf{I}^{\mathrm{im}}_{v,i}]
\in\mathbb{R}^{2K_f}$.
A residual MLP and a shared projection
$\mathcal{E}_f:\mathbb{R}^{2K_f}\rightarrow\mathbb{R}^{d}$
produce spectral features
$\mathbf{f}^h_{v,i}
=m^h_{v,i}\mathcal{E}_f(
\mathbf{s}^h_{v,i}
+\operatorname{MLP}_f(\mathbf{s}^h_{v,i}))
\in\mathbb{R}^{d}$.

\paragraph{Global spectrum aggregation.}
To summarize periodic patterns across the historical window,
we aggregate the point-level coefficients into a variable-level
spectrum. This supplies global information for comparing variables
and provides the historical spectrum used for future extrapolation.
With $n_v=\sum_{i=1}^{L_h}m^h_{v,i}$, we compute
$\bar{\mathbf{s}}_v
=\frac{1}{\max(n_v,1)}
\sum_{i=1}^{L_h}\mathbf{s}^h_{v,i}
=[\bar{\mathbf{R}}^{\mathrm{re}}_v;
\bar{\mathbf{I}}^{\mathrm{im}}_v]
\in\mathbb{R}^{2K_f}$,
and obtain the global frequency representation
$\mathbf f^g_v
=\mathcal E_f\!\left(
\bar{\mathbf s}_v+\operatorname{MLP}_f(\bar{\mathbf s}_v)
\right)$.

\paragraph{Future frequency representation extrapolation.}
Future queries have known timestamps but no observed values,
so their frequency features cannot be computed directly using
Eq.~\ref{eq:nudft}. Assuming that the historical spectral components remain relevant over the forecast horizon, we construct future frequency representations by rotating the aggregated historical spectrum according to the query timestamps.
For future position $j\in[L_p]$, define
$\psi_{v,j,k}=2\pi w_k\widehat{t}^p_{v,j}$ and rotate the
historical coefficients as:
\begin{equation}
\begin{bmatrix}
R^{p,\mathrm{re}}_{v,j,k}\\
I^{p,\mathrm{im}}_{v,j,k}
\end{bmatrix}
=
\begin{bmatrix}
\cos\psi_{v,j,k} & -\sin\psi_{v,j,k}\\
\sin\psi_{v,j,k} & \cos\psi_{v,j,k}
\end{bmatrix}
\begin{bmatrix}
\bar R^{\mathrm{re}}_{v,k}\\
\bar I^{\mathrm{im}}_{v,k}
\end{bmatrix},
\qquad k\in[K_f].
\label{eq:future_spectrum}
\end{equation}
Concatenating the rotated coefficients gives
$\mathbf{s}^p_{v,j}\in\mathbb{R}^{2K_f}$.
The same residual encoder produces
$\mathbf{f}^p_{v,j}
=\mathcal{E}_f(
\mathbf{s}^p_{v,j}
+\operatorname{MLP}_f(\mathbf{s}^p_{v,j}))
\in\mathbb{R}^{d}$.
If variable $v$ has no valid history, we set
$\mathbf{f}^g_v=\mathbf{0}$ and
$\mathbf{f}^p_{v,j}=\mathbf{0}$, so that future queries do not
introduce unsupported spectral evidence.

\subsection{Frequency-Guided Cross-Variable Interaction}
\label{sec:patch_interaction}


IMTS variables are observed asynchronously, and their dependencies may span multiple observations within local intervals. To support generalizable cross-variable interaction, we group observations into patches, select temporally nearby patches from other variables, and weight their contributions by frequency similarity.

\paragraph{Patch construction and pooling.}
To summarize local patterns for cross-variable interaction,
we pack valid historical and future tokens separately, in
timestamp order, into patches containing at most $B$ points. Let $\mathcal{P}_a$ denote patch $a$ and $v(a)$ its variable identity.
Within each patch, a learned query applies attention pooling over
valid temporal tokens, followed by a residual feed-forward layer,
to obtain $\mathbf{p}_a\in\mathbb{R}^{d}$.
The same pooling weights aggregate point-level frequency representations
into $\mathbf{f}_a\in\mathbb{R}^{d}$, which describes the patch's
local frequency patterns.

\paragraph{Asynchronous temporal neighbor selection.}
To identify related patches across asynchronously observed
variables, we compute distances between patch time intervals.
Let $[s_a,e_a]$ denote the valid time interval of patch $a$.
For each patch $a$, we compute
$\delta_{ab}=\max\{(s_a-e_b)_+,(s_b-e_a)_+\}$
and select the top-$K_a$ nearest patches from other variables:
$\mathcal{N}_{K_a}(a)
=\operatorname{TopK}_{b:\,v(b)\neq v(a)}(-\delta_{ab})$,
where $(x)_+=\max(x,0)$.
This distance is zero for overlapping intervals and equals the
gap between disjoint intervals.
The number of neighbors $K_a$ is bounded by $K_{\max}$ and the
available cross-variable candidates.

\paragraph{Frequency-aware routing and gated message injection.}
Being close in time does not imply similar value patterns, so giving all selected patches equal weight may introduce
unrelated information. We therefore use the frequency representations obtained through learnable spectrum projection and aggregation to determine their contributions. Patch-level representations $\mathbf{f}_a$ describe local patterns,
while variable-level representations $\mathbf{f}^g_{v(a)}$ summarize
broader historical patterns.
Their similarities jointly determine the aggregation weights:
\begin{equation}
\begin{aligned}
r_{ab}
&=
\frac{
\rho\cos(\mathbf{f}^g_{v(a)},\mathbf{f}^g_{v(b)})
+(1-\rho)\cos(\mathbf{f}_a,\mathbf{f}_b)
}{\tau_f},
\qquad
w_{ab}
=
\operatorname{softmax}_{b\in\mathcal{N}_{K_a}(a)}(r_{ab}),
\\[-0.5mm]
\mathbf{m}_a
&=
\tanh(\zeta)\mathbf{W}^{o}_m
\sum_{b\in\mathcal{N}_{K_a}(a)}
w_{ab}\mathbf{W}^{v}_m\mathbf{p}_b .
\end{aligned}
\label{eq:frequency_guided_interaction}
\end{equation}
Here, $\rho=\operatorname{sigmoid}(\bar{\rho})\in(0,1)$ balances
local and global similarities, $\tau_f>0$ is a learned temperature,
and $\zeta$ is a learned residual parameter initialized to zero.
Thus, the message $\mathbf{m}_a$ emphasizes temporal information
from patches with similar local and global frequency patterns.

To control how much of this message each future token receives,
we compute
$g_i=\operatorname{sigmoid}\!\left(
\mathbf{W}_g[\operatorname{LN}(\mathbf{h}_i);
\operatorname{LN}(\mathbf{m}_a)]\right)$
and update
$\mathbf{h}'_i
=\mathbf{h}_i+g_i\mathbf{W}_{\mathrm{inj}}\mathbf{m}_a$,
where $i\in\mathcal{P}^{p}_a$ and $g_i\in(0,1)$.



\subsection{Forecasting Objective and Pre-training}
\label{sec:pretraining}

For each valid future query $(v,j)$, we fuse its temporal and spectral representations as
$\mathbf{o}_{v,j}
=\operatorname{MLP}_{\mathrm{pred}}
([\mathbf{h}'^{\,p}_{v,j};\mathbf{f}^p_{v,j}])$
and predict
$(\widehat{u}^{p,q}_{v,j})_{q\in\mathcal{A}}
=\mathbf{W}_{\mathrm{head}}\mathbf{o}_{v,j}
+\mathbf{b}_{\mathrm{head}}$,
where $\mathcal{A}=\{q_1,\ldots,q_Q\}\subset(0,1)$ includes $0.5$.
Using historical statistics $\mu_v,s_v$, define
$u^p_{v,j}=\operatorname{asinh}((x^p_{v,j}-\mu_v)/s_v)$
and $e^q_{v,j}=u^p_{v,j}-\widehat{u}^{p,q}_{v,j}$.
For valid future targets
$\Omega=\{(v,j):v\in[V],\,j\in[L_p],\,m^p_{v,j}=1\}$,
the masked multi-quantile pinball objective is
\begin{equation}
    \mathcal{L}(\theta)
    =
    \frac{1}{|\Omega||\mathcal{A}|}
    \sum_{(v,j)\in\Omega}
    \sum_{q\in\mathcal{A}}
    \max\!\left\{
        q\,e^q_{v,j},
        (q-1)e^q_{v,j}
    \right\}.
    \label{eq:pinball}
\end{equation}
At inference, the median prediction is mapped back to the original scale as
$\widehat{x}^p_{v,j}
=\mu_v+s_v\sinh(\widehat{u}^{p,0.5}_{v,j})$.

\section{Experiments}

\subsection{Experiments Setup}

\paragraph{Pre-training and Downstream datasets}
We pretrain on four sources: GIFT-Eval Pretrain, GIFT-Eval~\citep{aksu2024gift} training dataset, the Chronos corpus, and our processed irregular corpus. There is no test data leakage. We collect eight open-source datasets spanning seven distinct domains for irregular time series forecasting. More details are showed in Appendix~\ref{app:pretraining_corpus} and ~\ref{app:downstream_benchmarks}.


\paragraph{Baselines}
To demonstrate the generalization ability of QiYao-I, we compare it with both irregular-input and regular-input \emph{time series foundation models}. 
To evaluate the few-shot learning ability of QiYao-I, we further compare it with a wide range of \emph{end-to-end irregular time series forecasting models}. More details are showed in Appendix~\ref{app:baselines}.


\paragraph{Implementation details}
For pre-training, we implement QiYao-I in PyTorch. The pre-training and zero-shot experiments are conducted on six NVIDIA A800 GPUs with 80GB memory. The few-shot experiments are conducted on four NVIDIA 3090 GPUs with 24GB memory. More details are shown in Appendix~\ref{app:model_settings} and ~\ref{app:evaluation_metrics}.



\begin{table*}[h]
\centering

\caption{
Average zero-shot forecasting results on downstream datasets.
Lower MSE and MAE values indicate better performance.
The best and second-best results are highlighted in
purple and blue, respectively.
}

\label{tab:zeroshot_results}

\renewcommand{\arraystretch}{1.15}
\setlength{\tabcolsep}{2pt}

\resizebox{\textwidth}{!}{%
\begin{tabular}{
@{}l|
cc|cc|cc|cc|cc|cc|cc|cc|
c@{}
}

\toprule


\rowcolor{headergray}

\textbf{Dataset}
&
\multicolumn{2}{c|}{\textbf{CESNET}}
&
\multicolumn{2}{c|}{\textbf{APTC}}
&
\multicolumn{2}{c|}{\textbf{ClusterTrace}}
&
\multicolumn{2}{c|}{\textbf{Seabirds}}
&
\multicolumn{2}{c|}{\textbf{Human Activity}}
&
\multicolumn{2}{c|}{\textbf{FNSPID}}
&
\multicolumn{2}{c|}{\textbf{GDELT}}
&
\multicolumn{2}{c|}{\textbf{USHCN}}
&
\cellcolor{white}
\multirow{3}{*}{
\shortstack{
\textbf{$1^{\mathrm{st}}$}\\
\textbf{Count}
}
}
\\

\cmidrule(lr){2-3}
\cmidrule(lr){4-5}
\cmidrule(lr){6-7}
\cmidrule(lr){8-9}
\cmidrule(lr){10-11}
\cmidrule(lr){12-13}
\cmidrule(lr){14-15}
\cmidrule(lr){16-17}


&
\multicolumn{2}{c|}{$(10^{-3})$}
&
\multicolumn{2}{c|}{$(10^{-5})$}
&
\multicolumn{2}{c|}{$(10^{-3})$}
&
\multicolumn{2}{c|}{$(10^{-6})$}
&
\multicolumn{2}{c|}{$(10^{-4})$}
&
\multicolumn{2}{c|}{$(10^{-5})$}
&
\multicolumn{2}{c|}{$(10^{-4})$}
&
\multicolumn{2}{c|}{$(10^{-3})$}
&
\\

\cmidrule(lr){2-3}
\cmidrule(lr){4-5}
\cmidrule(lr){6-7}
\cmidrule(lr){8-9}
\cmidrule(lr){10-11}
\cmidrule(lr){12-13}
\cmidrule(lr){14-15}
\cmidrule(lr){16-17}


\multirow{-2}{*}{
\cellcolor{white}
\textbf{Methods}
}
&
\textbf{MSE}
&
\textbf{MAE}
&
\textbf{MSE}
&
\textbf{MAE}
&
\textbf{MSE}
&
\textbf{MAE}
&
\textbf{MSE}
&
\textbf{MAE}
&
\textbf{MSE}
&
\textbf{MAE}
&
\textbf{MSE}
&
\textbf{MAE}
&
\textbf{MSE}
&
\textbf{MAE}
&
\textbf{MSE}
&
\textbf{MAE}
&
\\

\midrule


\rowcg
\multicolumn{18}{c}{
\textit{\textbf{Regular Time Series Foundation Models}}
}
\\[-1pt]

\midrule

\textbf{Toto-2}
&22.41&92.64
&\second{386.52}&\best{4595.26}
&\second{41.58}&\second{149.01}
&313.40&9030.82
&42.95&\best{379.01}
&\second{280.88}&\second{3027.27}
&198.90&\second{841.81}
&1064.11&437.90
&
\cellcolor{countred}\textbf{2}
\\

\textbf{Chronos-2}
&\best{22.00}&\second{90.94}
&401.85&4681.75
&45.99&157.43
&\second{295.50}&9384.00
&\second{42.43}&\second{386.46}
&311.00&3253.60
&198.44&843.06
&1123.39&\best{428.12}
&
\cellcolor{countred}\textbf{2}
\\

\textbf{TimesFM-2.5}
&22.38&91.79
&391.60&4618.95
&46.33&159.83
&358.00&9740.50
&55.64&434.63
&382.25&3601.35
&204.37&847.71
&1507.32&524.45
&
\cellcolor{countred}0
\\

\textbf{MOIRAI-2}
&22.45&91.02
&400.30&4669.20
&53.87&177.82
&306.00&\second{8895.50}
&48.06&403.01
&381.45&3592.50
&203.93&859.96
&1202.80&485.03
&
\cellcolor{countred}0
\\

\textbf{Sundial}
&22.50&92.06
&410.55&4735.05
&47.70&163.08
&341.00&11192.50
&44.57&400.52
&421.35&3887.70
&204.05&867.22
&\second{968.92}&\second{429.15}
&
\cellcolor{countred}0
\\

\textbf{TIME-MOE}
&22.86&95.63
&403.10&4692.85
&50.82&166.79
&522.00&15041.00
&57.61&468.22
&385.70&3704.25
&\best{196.51}&867.53
&992.84&468.72
&
\cellcolor{countred}\textbf{1}
\\

\textbf{Timer}
&33.97&137.50
&506.45&5357.75
&146.54&280.55
&55018.50&132425.00
&352.65&1125.54
&13941.95&23862.85
&242.93&1059.99
&1336.50&525.53
&
\cellcolor{countred}0
\\

\midrule


\rowcg
\multicolumn{18}{c}{
\textit{\textbf{Irregular Time Series Foundation Models}}
}
\\[-1pt]

\midrule

\textbf{MIRA}
&22.36&92.78
&410.21&4700.54
&48.86&166.95
&469.61&14857.72
&917.34&2667.64
&447.74&4528.26
&417.40&1321.87
&1138.82&493.80
&
\cellcolor{countred}0
\\

\textbf{QiYao-I}
&\second{22.01}&\best{90.24}
&\best{385.30}&\second{4597.09}
&\best{37.22}&\best{142.67}
&\best{165.67}&\best{7411.77}
&\best{41.45}&392.10
&\best{249.80}&\best{3016.12}
&\second{197.47}&\best{841.66}
&\best{907.77}&457.05
&
\cellcolor{countred}\textbf{11}
\\

\bottomrule
\end{tabular}%
}

\vspace{-5mm}
\end{table*}

\begin{table*}[!htbp]
\centering

\caption{
Average 10\% few-shot forecasting results of QiYao-I and full-shot results
of end-to-end irregular time-series models.
Lower MSE and MAE values indicate better performance.
The best and second-best results are highlighted in
purple and blue, respectively.
}
\label{tab:fewshot_results}

\renewcommand{\arraystretch}{1.15}
\setlength{\tabcolsep}{2pt}

\resizebox{\textwidth}{!}{%
\begin{tabular}{
@{}l|
cc|cc|cc|cc|cc|cc|cc|cc|
c@{}
}

\toprule


\rowcolor{headergray}

\textbf{Dataset}
&
\multicolumn{2}{c|}{\textbf{CESNET}}
&
\multicolumn{2}{c|}{\textbf{APTC}}
&
\multicolumn{2}{c|}{\textbf{ClusterTrace}}
&
\multicolumn{2}{c|}{\textbf{Seabirds}}
&
\multicolumn{2}{c|}{\textbf{Human Activity}}
&
\multicolumn{2}{c|}{\textbf{FNSPID}}
&
\multicolumn{2}{c|}{\textbf{GDELT}}
&
\multicolumn{2}{c|}{\textbf{USHCN}}
&
\cellcolor{white}
\multirow{3}{*}{
\shortstack{
\textbf{$1^{\mathrm{st}}$}\\
\textbf{Count}
}
}
\\

\cmidrule(lr){2-3}
\cmidrule(lr){4-5}
\cmidrule(lr){6-7}
\cmidrule(lr){8-9}
\cmidrule(lr){10-11}
\cmidrule(lr){12-13}
\cmidrule(lr){14-15}
\cmidrule(lr){16-17}


&
\multicolumn{2}{c|}{$(10^{-3})$}
&
\multicolumn{2}{c|}{$(10^{-5})$}
&
\multicolumn{2}{c|}{$(10^{-3})$}
&
\multicolumn{2}{c|}{$(10^{-6})$}
&
\multicolumn{2}{c|}{$(10^{-4})$}
&
\multicolumn{2}{c|}{$(10^{-5})$}
&
\multicolumn{2}{c|}{$(10^{-4})$}
&
\multicolumn{2}{c|}{$(10^{-3})$}
&
\\

\cmidrule(lr){2-3}
\cmidrule(lr){4-5}
\cmidrule(lr){6-7}
\cmidrule(lr){8-9}
\cmidrule(lr){10-11}
\cmidrule(lr){12-13}
\cmidrule(lr){14-15}
\cmidrule(lr){16-17}


\multirow{-2}{*}{
\cellcolor{white}
\textbf{Methods}
}
&
\textbf{MSE} & \textbf{MAE}
&
\textbf{MSE} & \textbf{MAE}
&
\textbf{MSE} & \textbf{MAE}
&
\textbf{MSE} & \textbf{MAE}
&
\textbf{MSE} & \textbf{MAE}
&
\textbf{MSE} & \textbf{MAE}
&
\textbf{MSE} & \textbf{MAE}
&
\textbf{MSE} & \textbf{MAE}
&
\\

\midrule


\rowcg
\multicolumn{18}{c}{
\textit{\textbf{Full-Shot End-to-End Irregular Models}}
}
\\[-1pt]

\midrule

\textbf{TFMixer}
& 23.45 & 101.45
& \second{385.00} & \second{4630.00}
& \second{41.40} & \second{150.15}
& 400.00 & 14600.00
& \best{36.50} & \best{362.00}
& \second{420.00} & \second{4340.00}
& 191.50 & 845.50
& \best{514.70} & \second{264.70}
&
\cellcolor{countred}\textbf{3}
\\

\textbf{ASTGI}
& \second{21.60} & \second{93.20}
& \second{385.00} & 4655.00
& 41.55 & 154.90
& \second{300.00} & \second{12500.00}
& 39.50 & 400.50
& 860.00 & 7845.00
& \best{188.00} & \second{840.50}
& 540.30 & 353.35
&
\cellcolor{countred}\textbf{1}
\\

\textbf{APN}
& 23.30 & 101.35
& \second{385.00} & 4640.00
& 43.60 & 153.60
& 1050.00 & 26100.00
& \second{37.00} & 375.00
& 1375.00 & 8145.00
& 190.00 & 855.50
& \second{518.60} & 333.35
&
\cellcolor{countred}0
\\

\textbf{KAFNet}
& 23.55 & 103.90
& \second{385.00} & \second{4630.00}
& 44.95 & 154.90
& 1000.00 & 27800.00
& 39.00 & 394.00
& \best{375.00} & 4665.00
& 189.00 & 860.00
& 640.55 & 414.85
&
\cellcolor{countred}\textbf{1}
\\

\textbf{Hi-Patch}
& 24.80 & 104.05
& \second{385.00} & \second{4630.00}
& 43.25 & 154.80
& 108900.00 & 281500.00
& 39.00 & 411.50
& 2295.00 & 10405.00
& 192.50 & 879.50
& 556.75 & 386.70
&
\cellcolor{countred}0
\\

\textbf{HyperIMTS}
& 28.40 & 128.10
& \second{385.00} & \second{4630.00}
& 43.95 & 158.05
& 450.00 & 15050.00
& 40.00 & 409.00
& 545.00 & 5385.00
& 189.50 & 849.50
& 692.05 & 486.35
&
\cellcolor{countred}0
\\

\textbf{GNeuralFlow}
& 23.45 & 103.75
& \second{385.00} & 4650.00
& 45.95 & 159.95
& 14950.00 & 100750.00
& 182.00 & 1052.50
& 1990.00 & 10640.00
& 190.50 & 860.00
& 628.80 & 410.55
&
\cellcolor{countred}0
\\

\textbf{tPatchGNN}
& 23.45 & 103.10
& 590.00 & 5910.00
& 45.40 & 155.70
& 1200.00 & 24350.00
& 37.50 & 387.50
& 915.00 & 7420.00
& 191.50 & \best{835.50}
& 665.60 & 437.40
&
\cellcolor{countred}\textbf{1}
\\

\textbf{GraFITi}
& 23.05 & 104.05
& \second{385.00} & \second{4630.00}
& 44.70 & 155.30
& 600.00 & 20100.00
& 38.00 & 387.50
& 1060.00 & 7580.00
& 199.50 & 919.00
& 638.80 & 428.15
&
\cellcolor{countred}0
\\

\textbf{Warpformer}
& 32.20 & 118.80
& 600.00 & 5970.00
& 91.55 & 229.35
& 142100.00 & 297150.00
& 124.00 & 784.00
& 3730.00 & 15865.00
& 1539.50 & 3253.00
& 544.95 & 390.90
&
\cellcolor{countred}0
\\

\textbf{CRU}
& 23.55 & 105.35
& \second{385.00} & \second{4630.00}
& 48.25 & 165.90
& 6050.00 & 54500.00
& 91.00 & 689.50
& 1040.00 & 7935.00
& 193.00 & 863.50
& 645.90 & 418.75
&
\cellcolor{countred}0
\\

\textbf{NeuralFlow}
& 23.45 & 103.45
& \second{385.00} & 4650.00
& 44.70 & 155.55
& 1550.00 & 29950.00
& 100.00 & 762.00
& 685.00 & 6150.00
& 190.00 & 853.00
& 571.45 & 372.15
&
\cellcolor{countred}0
\\

\textbf{mTAN}
& 23.35 & 103.75
& \second{385.00} & \second{4630.00}
& 49.90 & 168.70
& 2600.00 & 31450.00
& 61.00 & 569.00
& 845.00 & 7225.00
& 198.50 & 877.50
& 828.15 & 509.10
&
\cellcolor{countred}0
\\

\textbf{SeFT}
& 50.80 & 154.20
& 435.00 & 5025.00
& 99.65 & 244.70
& 8500.00 & 86000.00
& 166.00 & 1022.00
& 3670.00 & 16155.00
& 1576.50 & 3067.00
& 775.35 & 504.55
&
\cellcolor{countred}0
\\

\textbf{GRU-D}
& 23.30 & 103.20
& \second{385.00} & \second{4630.00}
& 45.45 & 155.70
& 28150.00 & 119950.00
& 48.50 & 492.00
& 1585.00 & 9960.00
& 189.00 & 851.50
& 664.40 & 426.45
&
\cellcolor{countred}0
\\

\midrule


\textbf{QiYao-I (10\%)}
& \best{21.59} & \best{87.49}
& \best{370.92} & \best{4507.52}
& \best{36.57} & \best{142.81}
& \best{171.39} & \best{7810.57}
& 39.52 & \second{368.79}
& 476.13 & \best{3916.12}
& \second{188.08} & 848.31
& 538.82 & \best{253.34}
&
\cellcolor{countred}\textbf{10}
\\

\bottomrule
\end{tabular}%
}

\vspace{-5mm}
\end{table*}

\subsection{Main Results}

\subsubsection{Compared to other foundation models}

To evaluate the generalization ability of QiYao-I on different irregular multivariate time series forecasting tasks, we compare it with various time series foundation models, including Toto2, Chronos2, TimesFM2.5, MOIRAI2, Sundial, TIME-MOE, and Timer, which are originally designed for regular time series inputs, as well as MIRA, which is originally designed for irregular
time series inputs. The experimental results are shown in Table ~\ref{tab:zeroshot_results}. Overall, QiYao-I achieves the best or second-best average results on most datasets, demonstrating cross-dataset generalization ability. In contrast, lower-ranked models show less stable performance across different irregular datasets. These results indicate that QiYao-I exhibits better performance and robustness under different data scales, prediction ranges, and time series distributions. Moreover, the evaluated learned interpolation method for regular-input models can lead to larger forecasting errors than linear interpolation, as shown in Appendix~\ref{interpolation method}.

\subsubsection{Compared to other end to end irregular models}

To evaluate the adaptation ability of QiYao-I, we compare its 10\% few-shot performance with 15 end-to-end irregular time series models trained on the full downstream training sets, including TFMixer, ASTGI, and etc.. As shown in Table~\ref{tab:fewshot_results}, QiYao-I achieves 10 best results among the 16 metrics, substantially more than all competing methods. It obtains the best MSE and MAE simultaneously on CESNET, APTC, ClusterTrace, and Seabirds, and also achieves the best MAE on FNSPID and USHCN. Moreover, it ranks second in MAE on Human Activity. These results demonstrate that, despite using only 10\% of the downstream training data, QiYao-I can effectively transfer its pretrained irregular temporal and cross-variable representations to heterogeneous target domains, providing stable few-shot performance across different sampling structure, and data scales. We further examine the contribution of pre-training
by comparing pre-trained and random initialization under same few-shot settings, with detailed results provided in
Appendix~\ref{app:initialization_comparison}.

\providecommand{\abcheck}{\textcolor{green!55!black}{\ding{51}}}
\providecommand{\abcross}{\textcolor{red!75!black}{\ding{55}}}

\begin{wraptable}{r}{0.55\textwidth}
\vspace{-3mm}
\setlength{\abovecaptionskip}{0pt}
\caption{
Ablation studies of QiYao-I. ``Full'' denotes dense interaction among all variables, while ``Aware'' denotes the proposed selective frequency-guided variable interaction. \abcheck\ and \abcross\ indicate enabled and disabled components, respectively. Bold values indicate the best results.
}
\label{tab:ablation_selected}
\centering
\footnotesize
\setlength{\tabcolsep}{2pt}
\renewcommand{\arraystretch}{1.35}

\resizebox{\linewidth}{!}{%
\begin{tabular}{
@{}c|
ccc|
cc|
cc|cc|cc@{}
}
\toprule

\multirow{3}{*}{\textbf{Variant}}
& \multicolumn{3}{c|}{\textbf{Temporal Modeling}}
& \multicolumn{2}{c|}{\textbf{Channel Modeling}}
& \multicolumn{6}{c}{\textbf{Forecasting Results}}
\\

\cmidrule(lr){2-4}
\cmidrule(lr){5-6}
\cmidrule(lr){7-12}

&
\multirow{2}{*}{\textbf{RoPE}}
&
\multirow{2}{*}{\textbf{\shortstack{CT-\\RoPE}}}
&
\multirow{2}{*}{\textbf{\shortstack{Manifold\\Bias}}}
&
\multirow{2}{*}{\textbf{Full}}
&
\multirow{2}{*}{\textbf{Aware}}
&
\multicolumn{2}{c|}{
    \textbf{\shortstack{FNSPID\\$(10^{-5})$}}
}
&
\multicolumn{2}{c|}{
    \textbf{\shortstack{Seabirds\\$(10^{-6})$}}
}
&
\multicolumn{2}{c}{
    \textbf{\shortstack{Human Activity\\$(10^{-4})$}}
}
\\

&
&
&
&
&
&
\textbf{MSE}
&
\textbf{MAE}
&
\textbf{MSE}
&
\textbf{MAE}
&
\textbf{MSE}
&
\textbf{MAE}
\\

\midrule

\rowcolor{gray!10}
\textbf{1}
& \abcheck
& \abcross
& \abcross
& \abcross
& \abcheck
& 305.27
& 3305.51
& 365.64
& 11787.29
& 50.65
& 449.02
\\

\textbf{2}
& \abcross
& \abcheck
& \abcross
& \abcross
& \abcheck
& 259.99
& 3245.83
& 576.66
& 15191.28
& 42.81
& 406.64
\\

\rowcolor{gray!10}
\textbf{3}
& \abcross
& \abcheck
& \abcheck
& \abcross
& \abcross
& 350.97
& 3651.22
& 558.11
& 14269.22
& 41.48
& 399.50
\\

\textbf{4}
& \abcross
& \abcheck
& \abcheck
& \abcheck
& \abcross
& 271.05
& 3062.03
& 177.10
& 8100.20
& 42.03
& 399.36
\\

\rowcolor{red!10}
\textbf{5}
& \abcross
& \abcheck
& \abcheck
& \abcross
& \abcheck
& \textbf{249.80}
& \textbf{3016.12}
& \textbf{165.67}
& \textbf{7411.77}
& \textbf{41.45}
& \textbf{392.10}
\\

\bottomrule
\end{tabular}%
}
\vspace{-5mm}
\end{wraptable}

\subsection{Model Analysis}
\textbf{Ablation Studies.}
To assess the proposed components, we conduct ablation studies on FNSPID, Seabirds, and Human Activity. As shown in Table~\ref{tab:ablation_selected}, adding the sampling-conditioned temporal manifold bias consistently outperforms standard RoPE and CT-RoPE, demonstrating the benefit of adapting temporal relations to local sampling structures. Removing cross-variable interaction or replacing frequency-guided selection with dense interaction also degrades performance, confirming the importance of selective information exchange across asynchronous variables. More analysis are shown in Appendix~\ref{app:bandwidth_intervention} and ~\ref{app:routing_intervention}.
\begin{wrapfigure}{r}{0.55\textwidth}
\vspace{-\intextsep}
\setlength{\abovecaptionskip}{2pt}
\setlength{\belowcaptionskip}{0pt}
\centering
\includegraphics[width=\linewidth]{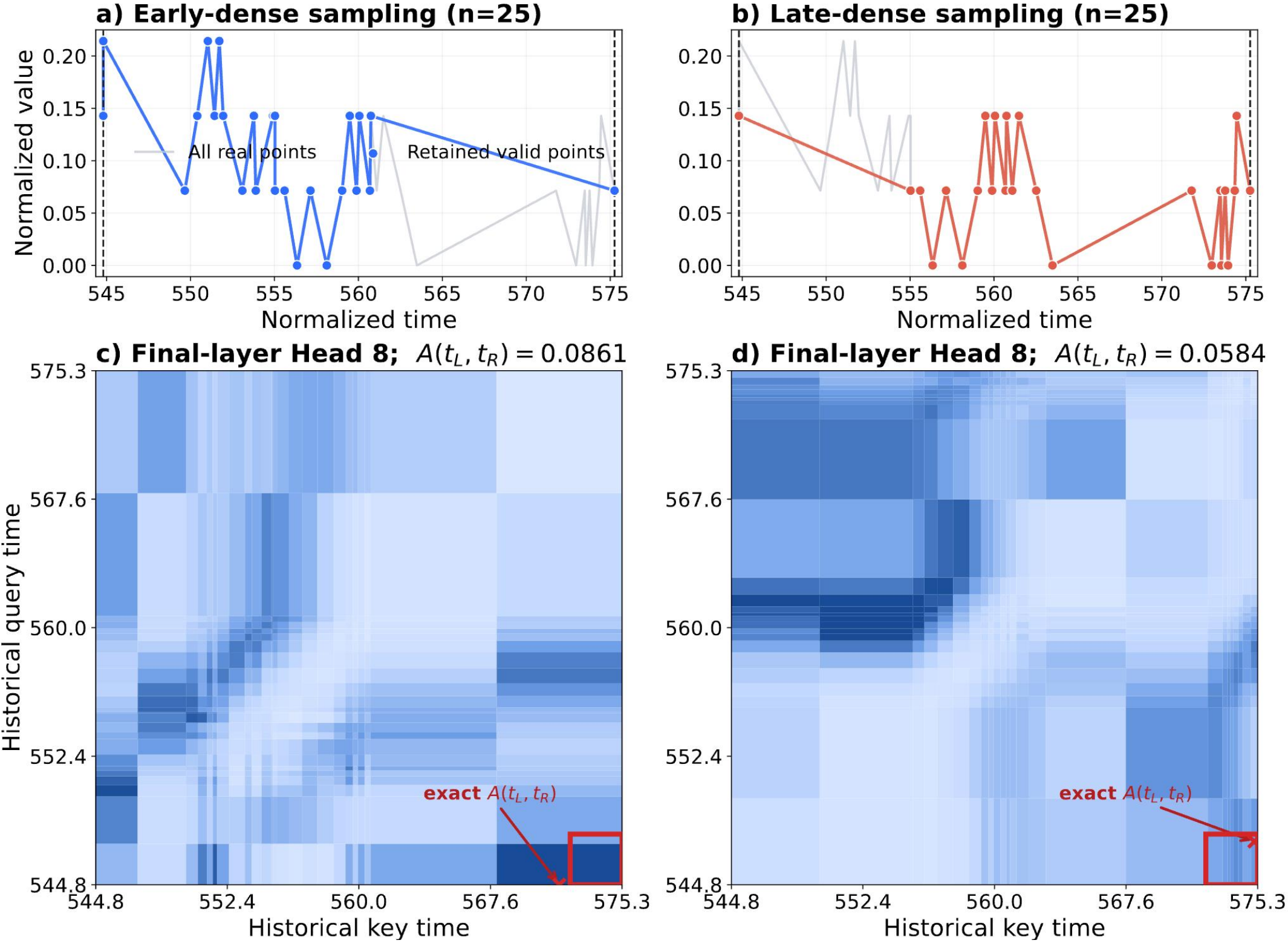}
\caption{Effect of local sampling structure on temporal attention.}
\label{fig:sampling_attention}
\vspace{-\intextsep}
\end{wrapfigure}

\textbf{Visualization of attention.}
To examine whether the proposed attention distinguishes same timestamp pairs under different local sampling structures, we construct two histories from the same ClusterTrace segment. They have same endpoint timestamps, the same temporal range, and the same number of valid observations, but their have different local sampling structure. As shown in Figure~\ref{fig:sampling_attention}, the same endpoint pair, $q=t_L=544.8$ and $k=t_R=575.3$, receives clearly different attention in the most sampling-responsive head. This result illustrates that the learned attention relation depends on the sampling structure rather than solely on the two endpoint timestamps.

More model analysis are shown in Appendix~\ref{effanalysis} and ~\ref{topk}.

\section{Conclusion}

We present QiYao-I, a manifold-based foundation model for irregular multivariate time series forecasting. It learns transferable representations through sampling-conditioned temporal manifold attention and frequency-guided patch interaction, capturing irregular temporal dependencies and asynchronous cross-variable relations. Extensive experiments demonstrate strong zero-shot generalization and few-shot adaptation. Our code is available at \url{https://anonymous.4open.science/r/QiYao-I-0053/}.
\newpage




\subsection*{AI use statement}

In this work, we used generative AI tools to assist with language polishing.
All AI-assisted text were carefully reviewed and verified by the authors.
The research questions, methodology, experimental design, analysis, and scientific claims were developed independently by the authors.
We take full responsibility for the final content of this work.

\subsection*{Ethics statement}

This work focuses on methodological research for irregular multivariate time series forecasting. Our experiments use existing pretraining datasets and downstream datasets, without involving human subjects or the intentional collection of personally identifiable information.
The proposed method is intended for general-purpose forecasting research, and we are not aware of any direct ethical risks arising from the methodology itself.




\subsection*{Reproducibility statement}

The model architecture of QiYao-I are described in Section~\ref{sec:method}, with experimental settings, dataset details, and implementation provided in Appendix~\ref{app:experimental_details}.
We report the configurations and evaluation protocols required to reproduce the main experiments.
The source code is available at \url{https://anonymous.4open.science/r/QiYao-I-0053/} and necessary scripts will be released.



\bibliographystyle{iclr2027_conference}
\bibliography{refs}

\newpage

\clearpage
\appendix

\section{Experimental Details}
\label{app:experimental_details}

\subsection{Pre-training Corpus and Sampling}
\label{app:pretraining_corpus}

\paragraph{Pretraining data sources.}
We pre-train QiYao-I on four sources: GIFT-Eval Pretrain, the training
portion of GIFT-Eval~\citep{aksu2024gift}, the Chronos corpus~\citep{ansari2025chronos}, and the
our collected irregular corpus from PyRegular~\citep{spinnato2025pyrregular}. The sampling probabilities of these sources are $0.30$, $0.25$, $0.20$, $0.25$, respectively. A streaming reader samples source records and
windows without loading the entire corpus into GPU memory. Samples from
different datasets appear in the same batch, but variables are never
assembled across unrelated records. For a native multivariate record, we
select at most twenty of its original variables; records containing fewer
variables retain their original dimensionality. A univariate sample follows
the same temporal and spectral path but has no cross-variable message.

\paragraph{Historical sampling augmentation.}
We retain original windows alongside four ways of modifying their
historical sampling pattern for MTS. Table~\ref{tab:app_augmentations} reports the
mixture used in pre-training. Point and block deletion update observation
masks without changing the future targets. Sparse subsampling retains a
subset of historical observations. Timestamp resampling draws new times
inside the historical window and linearly interpolates values only between
adjacent, originally valid historical observations. It neither interpolates
across an originally missing value nor uses future targets. Each variable
in a native multivariate record is augmented independently, producing
asynchronous views while preserving the original variable grouping.

\begin{table*}[h]
\centering
\caption{Historical sampling modes and their probabilities in the
pre-training stream. Original windows retain their native sampling.}
\label{tab:app_augmentations}
\small
\setlength{\tabcolsep}{5pt}
\begin{tabular}{@{}lcl@{}}
\toprule
Mode & Probability & Historical-view construction \\
\midrule
Original & $0.40$ & Preserve the original observations \\
Point deletion & $0.20$ & Independently remove valid points \\
Block deletion & $0.20$ & Remove one to three contiguous intervals \\
Timestamp resampling & $0.05$ & Resample times within valid historical intervals \\
Sparse subsampling & $0.15$ & Retain a strided subset of valid points \\
\bottomrule
\end{tabular}
\end{table*}

\subsection{Downstream Benchmarks}
\label{app:downstream_benchmarks}

We evaluate QiYao-I on eight open-source datasets from distinct domains: USHCN~\citep{menne2015long} from the climate domain, HumanActivity for activity, CESNET~\citep{koumar2025cesnet} for cybersecurity, APTC~\citep{DBLP:journals/eaai/Souza18} and ClusterTrace~\citep{weng2022mlaas} for engineering, FNSPID~\citep{DBLP:conf/kdd/DongFP24} for finance, GDELT~\citep{leetaru2013gdelt} for social events, and Seabirds~\citep{browning2018predicting} for nature. Dataset details are as followed. For these datasets, we select useful instances and variables, where we ensure that the time span and number of time points of all used samples are as similar as possible. Then we process them into a unified standard format. We summarize the statistics of these datasets in Table~\ref{tab:dataset_stats}. The collection is comprehensive, covering a wide range of domains, sampling patterns and missing rate. The “Samples” reported in Table~\ref{tab:dataset_stats} denotes the number of original continuous records following existing irregular works, rather than the number of forecasting instances used for training and evaluation. For datasets with sufficient records, we split the data at the record level. For datasets with only a few but sufficiently long records, such as CESNET, we perform a chronological split along the time axis and construct rolling forecasting windows within each split.
\begin{table*}[h]
\centering
\caption{The statistics of evaluation datasets.}
\label{tab:dataset_stats}
\resizebox{\textwidth}{!}{ 
\begin{tabular}{c|ccccccccc}
\toprule
Dataset       & Domain        & \# Variables & \# Samples & Avg \# obs. & Max Length & Missing Rate & Missing Pattern Complexity & Sampling Irregularity & Skewness \\
\midrule
HumanActivity & Activity      & 12           & 25         & 6593        & 304861.22  & 0.750        & 0.165                      & 0.396                 & 0.116    \\
USHCN         & Climate       & 5            & 1114       & 314         & 48.00      & 0.780        & 0.567                      & 0.999                 & 2.001    \\
CESNET        & Cybersecurity & 5            & 11         & 1700        & 1999.87    & 0.000        & 0.000                      & 0.637                 & 1.497    \\
ClusterTrace  & Engineering   & 3            & 3          & 5005        & 1525.58    & 0.057        & 0.119                      & 4.111                 & 0.795    \\
APTC          & Engineering   & 3            & 2109       & 399         & 1232.00    & 0.001        & 0.001                      & 0.000                 & -0.060   \\
FNSPID        & Finance       & 6            & 10         & 3495        & 5303.00    & 0.001        & 0.001                      & 0.606                 & 1.100    \\
GDELT         & Social        & 4            & 6          & 1684        & 1460.69    & 0.042        & 0.188                      & 1.279                 & 3.140    \\
Seabirds      & Nature        & 3            & 108        & 2442        & 10078.00   & 0.001        & 0.001                      & 0.283                 & 0.691   \\
\bottomrule
\end{tabular}
}
\end{table*}

Following existing settings, USHCN and APTC are randomly split into training, validation, and test sets with a ratio of 60\%/20\%/20\%, while Clustertrace, HumanActivity, CESNET, FNSPID, Seabirds, and GDELT use chronological splits within each instance before aggregating the corresponding segments. We use dataset-specific lookback and prediction lengths, which correspond to the real-time ranges of historical observations and future queries in each IMTS dataset, as shown in~\ref{tab:app_benchmark_settings}. The test loader follows the original sample-wise or time-wise split shown in the
table. We apply the same split and valid-target mask to every compared
model. The dataset-level normalization statistics are computed from the
training and validation portions, excluding the test portion. USHCN uses
z-score normalization; the other datasets use masked min--max
normalization. We evaluate all eligible test windows and exclude only
invalid future targets from metric calculations.

\begin{table*}[h]
\centering
\caption{Downstream evaluation configurations. $L_h$ is the historical
length and $L_p$ lists the two prediction lengths reported in the main
tables.}
\label{tab:app_benchmark_settings}
\small
\setlength{\tabcolsep}{5pt}
\begin{tabular}{@{}lccclc@{}}
\toprule
Dataset & $L_h$ & $L_p$ & Split & Normalization \\
\midrule
CESNET & 672 & 168, 336 & Time & Min--max \\
APTC & 400 & 50, 100 & Sample & Min--max \\
ClusterTrace & 168 & 24, 48 & Time & Min--max \\
Seabirds & 1440 & 360, 720 & Time & Min--max \\
Human Activity & 6000 & 200, 3000 & Time & Min--max \\
FNSPID & 730 & 180, 365 & Time & Min--max \\
GDELT & 365 & 30, 90 & Time & Min--max \\
USHCN & 24 & 1, 12 & Sample & Z-score \\
\bottomrule
\end{tabular}
\end{table*}

The following is dataset details:

\textbf{Human Activity}. This dataset consists of 12 irregularly measured 3D positional variables from sensors worn on the ankles, belts, and chests of five individuals performing various activities. We chunk the original time series into 25 IMTS. The first 6000 milliseconds are used as observed data to predict the sensor positions for the next 200/3000 milliseconds.

\textbf{USHCN}~\citep{menne2015long}. The USHCN dataset includes over 150 years of climate data from multiple U.S. stations, covering 5 climate variables. We focus on data from 1114 stations between 1996 and 2000, resulting in 1114 IMTS. Each instance uses data from the previous 24 months to predict the next 1/12 month’s climate conditions.

\textbf{CESNET}~\citep{koumar2025cesnet}. This dataset comprises network traffic data collected from the CESNET academic network, representing a large-scale collection of flow-level statistics. It contains 11 IMTS derived from millions of network flows, where each instance includes 5 features such as packet counts, byte counts, and inter-arrival times recorded over varying observation windows. Following standard protocol for network traffic analysis, the flows are processed into discrete intervals. We use the first 672 hours as the observed sequence to predict the traffic characteristics and flow types in the subsequent 168/336 hours.

\textbf{ClusterTrace}~\citep{DBLP:conf/nsdi/WengXYWWHLZLD22}. The ClusterTrace dataset provides a detailed trace of a production cluster containing multiple machines over a period of time. It captures complex workloads and resource utilization metrics, including CPU, memory, and disk usage for various tasks and containers. Due to the dynamic scheduling and heterogeneous nature of cloud workloads, the resource metrics are recorded as IMTS across thousands of unique job instances. Following common practices in cluster management research, the data is processed to focus on temporal resource demands. We select 3 IMTS and 3 variables for irregular time series forecasting. We use the resource usage data from the first 168 hours as the observed sequence to predict the workload intensity and resource requirements for the subsequent 24/48 hours.

\textbf{APTC}~\citep{DBLP:journals/eaai/Souza18}. This dataset is a large-scale collection of traffic speed data obtained from multiple dual-loop detectors on an urban expressway. After processed, it contains over 2109 IMTS instances, with each instance capturing 3 traffic-related variables. Due to detector malfunctions and data transmission issues, the dataset exhibits significant irregularity and missing values. We use the traffic observations from the previous 400 milliseconds as the observed sequence to predict the traffic conditions and flow patterns in the subsequent 50/100 milliseconds.

\textbf{FNSPID}~\citep{DBLP:conf/kdd/DongFP24}. The FNSPID dataset is a comprehensive Financial News and Stock Price Integration Dataset, combining stock price movements with corresponding financial news headlines. It covers thousands of stocks from major exchanges, featuring irregularly timed news events alongside stock price variables such as open, close, and volume. The alignment of sparse news signals with continuous price data results in multiple multi-modal IMTS. We select 10 IMTS and 6 variables for the irregular time series forecasting. To facilitate stock trend and volatility forecasting, each instance is organized into windows of sequential market activities. We use the integrated data from the previous 730 days as observed data to predict the stock price fluctuations and market sentiment for the following 180/365 days.

\textbf{GDELT}~\citep{leetaru2013gdelt}. This dataset is sourced from the Global Database of Events, Language, and Tone, which monitors world broadcast, print, and web news in multiple languages. It captures physical activities and socio-political events globally, resulting in a massive collection of IMTS with varying temporal resolutions based on event occurrences. The dataset includes multiple variables such as the Goldstein scale, average tone, and event counts across thousands of distinct themes and locations. To adapt it for the irregular time series forecasting, we select 6 IMTS instances and 4 variables. We use the event dynamics from the previous 365 days as the observed data to predict the event intensity and regional stability in the subsequent 30/90 days.

\textbf{Seabirds}~\citep{browning2018predicting}. This dataset tracks the movement and behavioral patterns of seabirds using high-resolution GPS and tri-axial accelerometer sensors. It contains IMTS representing multiple distinct movement variables, such as altitude, speed, and body orientation, collected from multiple species during foraging trips. Given the irregular nature of wildlife tracking due to varying signal quality and nesting cycles, the data is segmented into multiple IMTS instances, each spanning a specific duration of flight or diving activity. We select 108 IMTS and 3 variables for the irregular time series forecasting. We use the sensor measurements from the first 1440 minutes as the observed data to predict the bird's trajectory and behavioral states in the subsequent 360/720 minutes.

\subsection{Baselines and Input Adaptation}
\label{app:baselines}

\paragraph{Foundation models.}
The zero-shot comparison includes Toto-2~\citep{khwaja2026toto}, Chronos-2~\citep{ansari2025chronos}, TimesFM-2.5~\citep{das2023decoder},
Moirai-2~\citep{liu2025moirai}, Sundial~\citep{liu2025sundial}, Time-MoE~\citep{shi2025time}, and Timer~\citep{liu2024timer}, which accept regularly sampled inputs, as well as MIRA~\citep{li2026mira}, which supports irregular timestamps. For regular-input models, irregular historical observations are converted to a regular input grid by interpolation, as shown in~\ref{alg:linear_resampling}. QiYao-I and MIRA receive irregular
timestamps and observation masks directly. We use each model's median
forecast as its point prediction when probabilistic outputs are available. Every model is scored against the same original valid future targets.

\begin{algorithm}[h]
\caption{Linear Resampling for Irregular Historical Observations}
\label{alg:linear_resampling}
\begin{algorithmic}[1]
\REQUIRE Irregular time points $T$, observations $X$, mask $M$, number of regular samples $L$
\ENSURE Regularly sampled sequence $X^{\mathrm{reg}}$

\FOR{each sample $i$}
    \STATE Find the valid historical time range according to $M_i$
    \STATE Generate a regular grid $T_i^{\mathrm{reg}}$ within this range
    \FOR{each variable $d$}
        \STATE Select valid observations according to $M_{i,d}$
        \IF{there is no valid observation}
            \STATE Set $X_{i,d}^{\mathrm{reg}} \leftarrow 0$
        \ELSIF{there is only one valid observation}
            \STATE Set $X_{i,d}^{\mathrm{reg}}$ as a constant sequence
        \ELSE
            \STATE Sort observations by time
            \STATE Average values with duplicated timestamps
            \STATE Linearly interpolate observations onto $T_i^{\mathrm{reg}}$
        \ENDIF
    \ENDFOR
\ENDFOR

\RETURN $X^{\mathrm{reg}}$
\end{algorithmic}
\end{algorithm}

\paragraph{End-to-end irregular models.}
The few-shot comparison includes TFMixer~\citep{qiu2026bridging}, ASTGI~\citep{DBLP:journals/corr/abs-2509-23313}, APN~\citep{DBLP:journals/corr/abs-2505-11250}, KAFNet~\citep{zhou2026revitalizing},
Hi-Patch~\citep{DBLP:conf/icml/LuoZ0025}, HyperIMTS ~\citep{li2025hyperimts}, GNeuralFlow~\citep{mercatali2024graph}, tPatchGNN~\cite{DBLP:conf/icml/ZhangYL0024}, GraFITi~\cite{DBLP:conf/aaai/YalavarthiMSABJ24}, Warpformer~\cite{DBLP:conf/kdd/ZhangZCBL23},
CRU~\cite{DBLP:conf/icml/SchirmerELR22}, NeuralFlow~\cite{DBLP:conf/nips/BilosSRJG21}, mTAN~\cite{DBLP:conf/iclr/ShuklaM21}, SeFT~\cite{DBLP:conf/icml/HornMBRB20}, and GRU-D~\cite{DBLP:journals/corr/ChePCSL16}. As indicated in the main
table, these baselines use their full training splits, whereas QiYao-I
is adapted using only $10\%$ of the corresponding training data.
This difference in training-data access is retained explicitly in
the table caption and should be considered when interpreting rankings.

\subsection{Model and Optimization Settings}
\label{app:model_settings}

Table~\ref{tab:app_model_config} lists the configuration used by the
reported QiYao-I checkpoint. The sampling-measure kernel width
$\kappa$ controls how each anchor counts observations; it is distinct
from the learned effective bandwidth of the manifold attention bias.
During pre-training, the number of time-nearest cross-variable patches
is sampled from $1$ to $16$ when interaction is active. History length is set 576 and prediction length is set 192.

\begin{table*}[h]
\centering
\caption{QiYao-I architecture used in the main experiments.}
\label{tab:app_model_config}
\small
\setlength{\tabcolsep}{7pt}
\begin{tabular}{@{}lrlr@{}}
\toprule
Configuration & Value & Configuration & Value \\
\midrule
Transformer layers & 3 & Hidden dimension & 256 \\
Attention heads & 8 & Feed-forward dimension & 1024 \\
Time embedding dimension & 32 & Manifold dimension / head & 16 \\
Sampling anchors $A$ & 9 & Sampling window $W$ & 32 \\
Descriptor kernel width $\kappa$ & 0.2 & Time reference $R$ & 576 \\
Learnable frequencies & 16 & Patch size & 8 \\
Maximum native variables & 20 & Forecast quantiles & 21 \\
Maximum neighbors $K$ & 16 & Dropout & 0.1 \\
\bottomrule
\end{tabular}
\end{table*}

\paragraph{Pre-training stages.}
QiYao-I is first trained on univariate windows to learn stable sampling-conditioned temporal manifold attention. The cross-variable module remains in the architecture but has no eligible neighbors for these samples. Then, the multivariate stage starts
from the best univariate checkpoint, initializes a new AdamW
optimizer. This stage uses native multivariate records and 
also include univariate samples for joint training. 
Both stages use FP32. The univariate stage uses six NVIDIA A800 GPUs, and the multivariate stage also uses six NVIDIA A800 GPUs.
Both use AdamW with a peak learning rate of $6\times10^{-5}$, 1000
warm-up steps, cosine decay to $6\times10^{-6}$, weight decay $0.01$,
and gradient clipping at norm $1.0$. The training seed is 260914.

\paragraph{Ten-percent few-shot adaptation.}
For each downstream dataset and prediction length, we start from the
same pre-trained checkpoint and train on a fixed $10\%$ subset of its
training split. USHCN retains z-score normalization during both adaptation
and evaluation.

\subsection{Evaluation Metrics and Aggregation}
\label{app:evaluation_metrics}

All metrics use the median forecast after the model's inverse value
transformation. For dataset variable $v$, let $\mathcal{Q}_v$ be its
valid future targets across test samples, and let $n_v=|\mathcal{Q}_v|$.
We first compute variable-level errors
\begin{equation}
E^{\mathrm{MSE}}_v=
\frac{1}{n_v}\sum_{j\in\mathcal{Q}_v}(y_{v,j}-\widehat y_{v,j})^2,
\qquad
E^{\mathrm{MAE}}_v=
\frac{1}{n_v}\sum_{j\in\mathcal{Q}_v}|y_{v,j}-\widehat y_{v,j}|.
\label{eq:app_variable_errors}
\end{equation}
Only variables with $n_v>0$ enter the reported macro average:
$\mathrm{Macro\text{-}MSE}=|\mathcal{V}|^{-1}
\sum_{v\in\mathcal{V}}E^{\mathrm{MSE}}_v$ and
$\mathrm{Macro\text{-}MAE}=|\mathcal{V}|^{-1}
\sum_{v\in\mathcal{V}}E^{\mathrm{MAE}}_v$, where
$\mathcal{V}=\{v:n_v>0\}$. For each dataset, the \textsc{Avg} row is
the arithmetic mean of its two reported prediction lengths,
calculated separately for MSE and MAE.

\newpage
\section{Additional model analysis}
\label{app:additional_model_analysis}

\subsection{Efficiency analysis}
\label{effanalysis}
\begin{figure}[h]
\centering
\includegraphics[width=\linewidth]{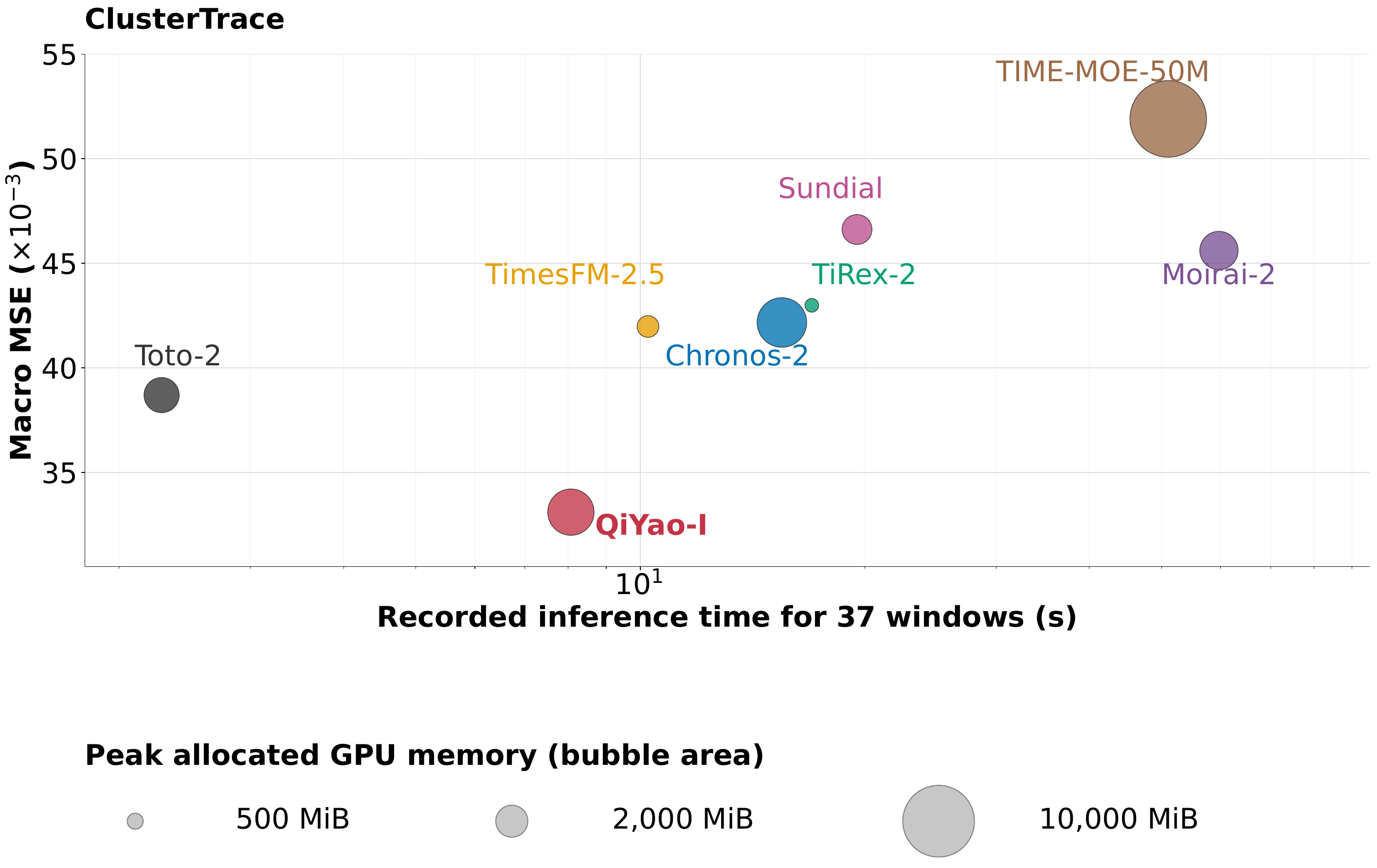}
\caption{Efficiency analysis.}
\label{eff}
\end{figure}

We evaluate inference efficiency on ClusterTrace using 37 test windows in 10 batches of up to four. Figure~\ref{eff} compares inference time, peak allocated GPU memory (bubble area), and forecasting error. QiYao-I achieves the lowest macro MSE, with 8.07\,s of recorded inference time and 4.08\,GiB of peak GPU allocation. Toto-2 is faster and uses less memory, but its MSE is higher; all other profiled baselines also yield larger errors.

\subsection{Analysis of TopK selection}
\label{topk}
\begin{figure}[h]
\centering
\includegraphics[width=\linewidth]{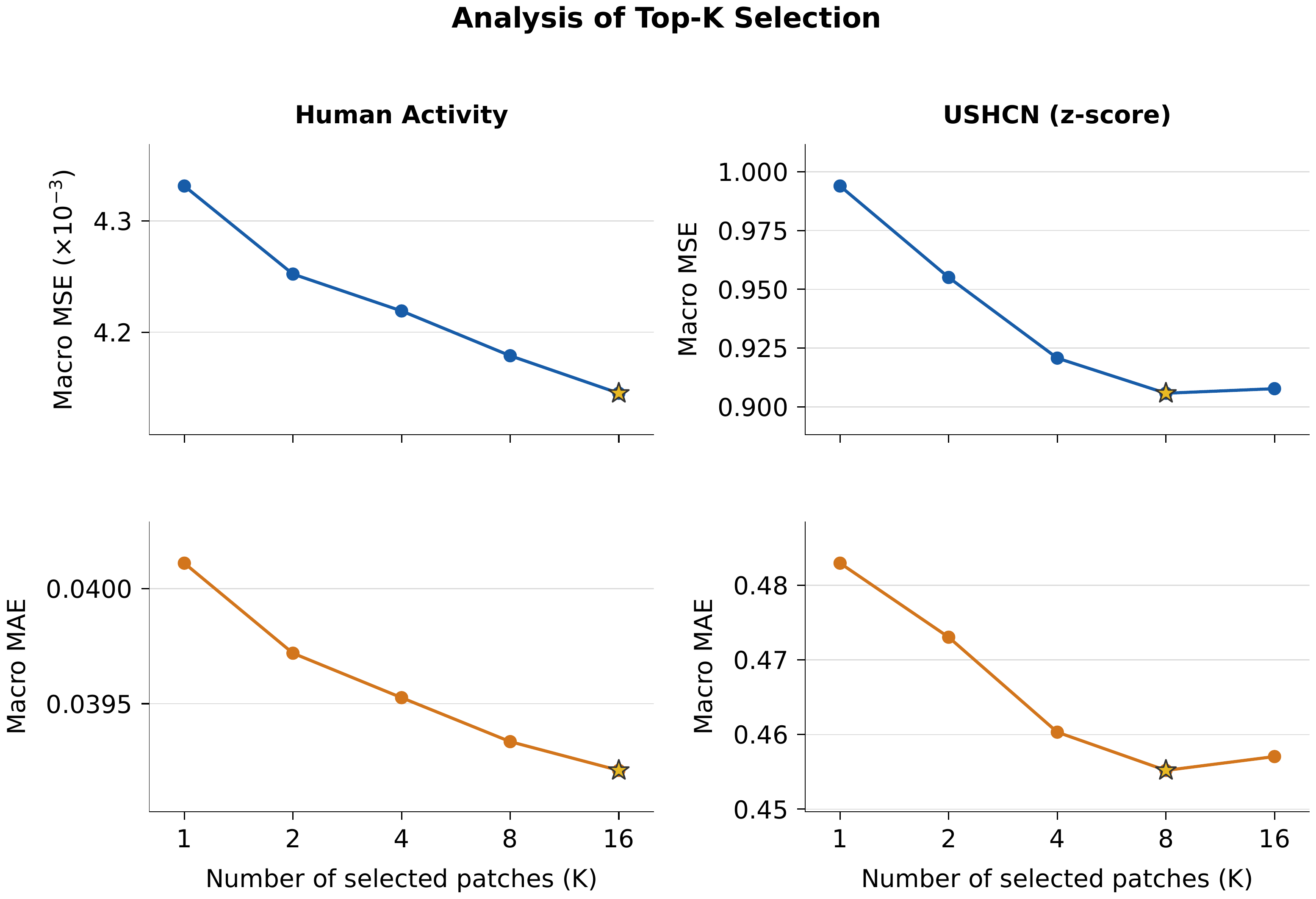}
\caption{TopK analysis.}
\label{fig:topk_sensitivity}
\end{figure}

Figure~\ref{fig:topk_sensitivity} examines the number of cross-variable patches selected for interaction. We vary $K$ at inference while keeping the pre-trained model fixed, and report the average macro MSE and MAE over two prediction lengths for each dataset. On Human Activity, both errors decrease as $K$ increases from 1 to 16, suggesting that additional neighboring patches provide useful information. On USHCN, both metrics improve up to $K=8$ but rise slightly at $K=16$. Thus, a larger neighbor set is not always beneficial: its optimal size depends on how much relevant information the selected patches provide. These results support using selective rather than unrestricted cross-variable interaction.

\subsection{Analysis of interpolation method}
\label{interpolation method}
Following prior irregular time series work, we use linear interpolation~\citep{li2026mira} as a simple and effective way to convert irregular observations into regular inputs for regular-input models. We construct a regular grid over the original time range and linearly estimate each variable from neighboring observations. This is used for all regular-input baselines and does not require any extra training. In contrast, learned imputation would introduce another trained modules and extra hyperparameters, making the comparison less direct. Nevertheless, we additionally evaluated Chronos-2, Toto2, and MOIRAI2 with irregular inputs first interpolated by BayOTIDE~\citep{fang2023bayotide}, a SOTA interpolation method. The details are shown in the Table~\ref{tab:interpolation_comparison}.QiYao-I still performs well.

\begin{table*}[t]
\centering
\caption{Zero-shot forecasting results with linear and BayOTIDE interpolation.
Results are averaged over two prediction lengths: CESNET (168, 336),
FNSPID (180, 365), and Human Activity (200, 3000).
All baseline point forecasts use the median ($q=0.5$).
QiYao-I directly models irregular observations.
Lower values are better; bold indicates the best result.}
\label{tab:interpolation_comparison}
\renewcommand{\arraystretch}{1.12}
\setlength{\tabcolsep}{7pt}
\begin{tabular}{l|cc|cc|cc}
\toprule
\multirow{2}{*}{\textbf{Method}}
& \multicolumn{2}{c|}{\textbf{CESNET} ($10^{-3}$)}
& \multicolumn{2}{c|}{\textbf{FNSPID} ($10^{-5}$)}
& \multicolumn{2}{c}{\textbf{Human Activity} ($10^{-4}$)} \\
& MSE & MAE & MSE & MAE & MSE & MAE \\
\midrule
\multicolumn{7}{c}{\textit{Linear interpolation}} \\
\midrule
Chronos-2 & \textbf{22.00} & 90.94 & 311.00 & 3253.60 & 42.43 & 386.46 \\
Toto-2    & 22.41 & 92.64 & 280.88 & 3027.27 & 42.95 & \textbf{379.01} \\
Moirai-2  & 22.45 & 91.02 & 381.45 & 3592.50 & 48.06 & 403.01 \\
\midrule
\multicolumn{7}{c}{\textit{BayOTIDE interpolation}} \\
\midrule
Chronos-2 & 22.67 & 91.96 & 1377.25 & 8382.05 & 240.41 & 1242.97 \\
Toto-2    & 28.12 & 99.81 & 2877.98 & 12947.19 & 203.70 & 1162.71 \\
Moirai-2  & 22.28 & 91.84 & 783.18 & 6950.05 & 114.40 & 874.15 \\
\midrule
\multicolumn{7}{c}{\textit{Direct irregular input}} \\
\midrule
\textbf{QiYao-I}
& 22.01 & \textbf{90.24}
& \textbf{249.80} & \textbf{3016.12}
& \textbf{41.45} & 392.10 \\
\bottomrule
\end{tabular}
\end{table*}

\subsection{Sampling-Conditioned Bandwidth}
\label{app:bandwidth_intervention}

\paragraph{Experimental setup.}
We evaluate ClusterTrace with prediction lengths 24 and 48,
and USHCN with prediction lengths 1 and 12.
We compare three settings:
\textit{Full}, which retains the learned sampling-conditioned bias;
\textit{Fixed}, which sets $r_u^{\ell,r}=1$ while preserving
the learned manifold distances and the parameters
$\beta_{\ell,r}$ and $\eta_{\ell,r}$; and
\textit{No bias}, which sets $b_{uw}^{\ell,r}=0$.
All remaining model parameters and operations are unchanged.

\begin{table}[t]
\centering
\caption{
Bandwidth interventions.
Original denotes the complete test set.
AVG gives the equally weighted average over the two prediction
lengths. Lower values are better.
}
\label{tab:bandwidth_intervention}
\renewcommand{\arraystretch}{1.12}
\setlength{\tabcolsep}{4pt}
\resizebox{\linewidth}{!}{%
\begin{tabular}{lll|cc|cc|cc}
\toprule
Dataset & Pred. & Input
& \multicolumn{2}{c|}{Full}
& \multicolumn{2}{c|}{Fixed ($r=1$)}
& \multicolumn{2}{c}{No bias ($b=0$)} \\
& & & MSE & MAE & MSE & MAE & MSE & MAE \\
\midrule
ClusterTrace & 24 & Original
& 0.031973 & 0.132937
& 0.033591 & 0.136933
& 0.040467 & 0.151741 \\
& 48 & Original
& 0.042463 & 0.152406
& 0.046573 & 0.159275
& 0.044720 & 0.159278 \\
& AVG & Original
& 0.037218 & 0.142671
& 0.040082 & 0.148104
& 0.042594 & 0.155509 \\
\midrule
USHCN & 1 & Original
& 0.928796 & 0.459120
& 0.940851 & 0.463556
& 0.949627 & 0.465101 \\
& 12 & Original
& 0.886734 & 0.454973
& 0.888887 & 0.455911
& 0.895354 & 0.468892 \\
& AVG & Original
& 0.907765 & 0.457046
& 0.914861 & 0.459734
& 0.922491 & 0.466997 \\
\bottomrule
\end{tabular}%
}
\end{table}

\paragraph{Results.}
As shown in Table~\ref{tab:bandwidth_intervention}, the full model
achieves lower MSE and MAE than both intervention settings across
the evaluated prediction lengths on ClusterTrace and USHCN.
Compared with fixed bandwidths, sampling-conditioned bandwidths
provide larger improvements on ClusterTrace and smaller but
consistent improvements on USHCN.
Removing the manifold bias also increases forecasting errors
on both datasets.
Under this frozen-checkpoint evaluation, these results suggest
that the learned temporal bias contributes to forecasting and
that adapting its bandwidth to local sampling structures provides
additional benefits.

\paragraph{Response to local sampling structure.}
We additionally examine the shared left-to-right endpoint pair
in every eligible variable and selected window, covering all
three temporal layers and eight attention heads.
For effective bandwidth
$B_{uw}^{\ell,r}
=\beta_{\ell,r}\sqrt{r_u^{\ell,r}r_w^{\ell,r}}$,
we measure the absolute relative change
$100|B_{uw}^{\ell,r,\mathrm{late}}/
B_{uw}^{\ell,r,\mathrm{early}}-1|$.
We also measure the absolute change in the endpoint bias.
Table~\ref{tab:sampling_response_statistics} averages these quantities
over all eligible variable-window and layer-head combinations.

\begin{table}[t]
\centering
\caption{
Endpoint responses to different local sampling structures.
The paired inputs have same endpoints and point counts.
$\Delta B$ is the mean absolute relative bandwidth change;
$\Delta b$ is the mean absolute change in the adaptive bias.
The last column reports the maximum absolute change in the
fixed-bandwidth bias.
}
\label{tab:sampling_response_statistics}
\renewcommand{\arraystretch}{1.12}
\setlength{\tabcolsep}{5pt}
\resizebox{\linewidth}{!}{%
\begin{tabular}{lrrrrrr}
\toprule
Dataset & Pred. & Windows & Variable-window pairs
& $\Delta B$ (\%) & $\Delta b$ & Fixed-bias change \\
\midrule
ClusterTrace & 24 & 37  & 111 & 5.51 & 2382.79 & 0 \\
ClusterTrace & 48 & 19  & 57  & 5.39 & 2456.90 & 0 \\
USHCN        & 1  & 128 & 640 & 1.96 & 1305.74 & 0 \\
USHCN        & 12 & 128 & 640 & 1.98 & 1315.91 & 0 \\
\bottomrule
\end{tabular}%
}
\end{table}

The fixed-bandwidth endpoint bias remains unchanged, whereas
the adaptive bandwidth and bias respond to the changed sampling
structure in all 24 layer-head combinations.
This confirms that the mechanism distinguishes sampling patterns
beyond the endpoint interval and observation count.

\subsection{Frequency-Based Routing with Fixed Neighbors}
\label{app:routing_intervention}

\paragraph{Experimental setup.}
To isolate the contribution of frequency-based aggregation weights,
we evaluate Human Activity with prediction lengths 200 and 3000,
and FNSPID with prediction lengths 180 and 365.
For each input batch, we verify that all compared settings use
same temporal Top-$K$ neighbor indices and validity masks.
We compare \textit{Local + Global}, \textit{Uniform},
\textit{Local only}, and \textit{Global only} routing.

\begin{table}[t]
\centering
\caption{
Frozen-checkpoint routing interventions with identical temporal
neighbors. AVG denotes the equally weighted average over the two
prediction lengths. All values are unscaled.
}
\label{tab:routing_intervention}
\renewcommand{\arraystretch}{1.12}
\setlength{\tabcolsep}{3pt}
\resizebox{\linewidth}{!}{%
\begin{tabular}{ll|cc|cc|cc|cc}
\toprule
Dataset & Pred.
& \multicolumn{2}{c|}{Local + Global}
& \multicolumn{2}{c|}{Uniform}
& \multicolumn{2}{c|}{Local only}
& \multicolumn{2}{c}{Global only} \\
& & MSE & MAE & MSE & MAE & MSE & MAE & MSE & MAE \\
\midrule
Human Activity & 200
& 0.00291967 & 0.03096386
& 0.00293623 & 0.03101229
& 0.00291467 & 0.03093724
& 0.00293426 & 0.03104303 \\
& 3000
& 0.00537056 & 0.04745604
& 0.00538634 & 0.04739915
& 0.00537137 & 0.04746498
& 0.00536983 & 0.04744481 \\
& AVG
& 0.00414511 & 0.03920995
& 0.00416129 & 0.03920572
& 0.00414302 & 0.03920111
& 0.00415204 & 0.03924392 \\
\midrule
FNSPID & 180
& 0.00148230 & 0.02358637
& 0.00150630 & 0.02361973
& 0.00148254 & 0.02358576
& 0.00148155 & 0.02358435 \\
& 365
& 0.00351367 & 0.03673599
& 0.00354458 & 0.03687244
& 0.00351467 & 0.03674194
& 0.00351207 & 0.03672738 \\
& AVG
& 0.00249798 & 0.03016118
& 0.00252544 & 0.03024608
& 0.00249860 & 0.03016385
& 0.00249681 & 0.03015587 \\
\bottomrule
\end{tabular}%
}
\end{table}

\paragraph{Results.}
As shown in Table~\ref{tab:routing_intervention}, combined frequency
routing reduces average MSE relative to uniform aggregation by
0.39\% on Human Activity and 1.09\% on FNSPID.
The corresponding MAE is nearly unchanged on Human Activity
and decreases by 0.28\% on FNSPID.
However, local-only routing yields slightly better average results
on Human Activity, while global-only routing performs slightly
better on FNSPID.
These findings support modest, task-dependent benefits of frequency
weighting over equal weights.

\subsection{Pre-trained versus Random Initialization}
\label{app:initialization_comparison}

\paragraph{Experimental setup.}
We compare pre-trained and random initialization of the same complete
architecture on CESNET (168/336), ClusterTrace (24/48), and USHCN (1/12).
For each prediction length, both initializations use the same fixed
10\% training subset across three optimization seeds.

We retain the existing preprocessing protocol, with normalization
statistics estimated from the training and validation splits,
excluding the test split.
Thus, 10\% refers to the training windows used for gradient updates.
USHCN uses z-score normalization, while the other datasets use
min--max normalization.


\begin{table}[t]
\centering
\caption{
Comparison of pre-trained and random initialization using the same
fixed 10\% training subsets and training settings.
Results are reported as mean $\pm$ sample standard deviation over
three optimization seeds.
AVG first averages the two prediction lengths within each
repetition and then summarizes the three repetitions.
The lower mean in each comparison is shown in bold.
}
\label{tab:initialization_comparison}
\renewcommand{\arraystretch}{1.15}
\setlength{\tabcolsep}{4pt}
\resizebox{\linewidth}{!}{%
\begin{tabular}{llcc}
\toprule
Dataset / Pred. & Initialization & Macro-MSE & Macro-MAE \\
\midrule

CESNET / 168
& Pre-trained
& $\mathbf{0.021285} \pm 0.000063$
& $\mathbf{0.087062} \pm 0.000023$ \\
& Random
& $0.021836 \pm 0.000754$
& $0.088425 \pm 0.002280$ \\

CESNET / 336
& Pre-trained
& $\mathbf{0.021522} \pm 0.000275$
& $\mathbf{0.087548} \pm 0.000318$ \\
& Random
& $0.022771 \pm 0.000503$
& $0.090517 \pm 0.000634$ \\

\textbf{CESNET / AVG}
& Pre-trained
& $\mathbf{0.021403} \pm 0.000162$
& $\mathbf{0.087305} \pm 0.000164$ \\
& Random
& $0.022304 \pm 0.000353$
& $0.089471 \pm 0.001335$ \\

\midrule

ClusterTrace / 24
& Pre-trained
& $\mathbf{0.032888} \pm 0.000656$
& $\mathbf{0.133225} \pm 0.001104$ \\
& Random
& $0.039602 \pm 0.000496$
& $0.149163 \pm 0.000718$ \\

ClusterTrace / 48
& Pre-trained
& $\mathbf{0.041636} \pm 0.001027$
& $\mathbf{0.153120} \pm 0.000885$ \\
& Random
& $0.043186 \pm 0.001579$
& $0.155469 \pm 0.002915$ \\

\textbf{ClusterTrace / AVG}
& Pre-trained
& $\mathbf{0.037262} \pm 0.000633$
& $\mathbf{0.143173} \pm 0.000343$ \\
& Random
& $0.041394 \pm 0.000562$
& $0.152316 \pm 0.001213$ \\

\midrule





USHCN / 1
& Pre-trained
& $0.564125 \pm 0.003276$
& $0.262441 \pm 0.002382$ \\
& Random
& $\mathbf{0.562252} \pm 0.003004$
& $\mathbf{0.261325} \pm 0.002290$ \\

USHCN / 12
& Pre-trained
& $\mathbf{0.517573} \pm 0.000536$
& $\mathbf{0.246203} \pm 0.000415$ \\
& Random
& $0.527935 \pm 0.010005$
& $0.246396 \pm 0.001192$ \\

\textbf{USHCN / AVG}
& Pre-trained
& $\mathbf{0.540849} \pm 0.001812$
& $0.254322 \pm 0.001002$ \\
& Random
& $0.545094 \pm 0.004568$
& $\mathbf{0.253861} \pm 0.000550$ \\

\bottomrule
\end{tabular}%
}
\end{table}

\paragraph{Results.}
As shown in Table~\ref{tab:initialization_comparison}, pre-trained
initialization achieves lower mean MSE and MAE at both prediction
lengths on CESNET and ClusterTrace.
Compared with random initialization, it reduces average MSE by
4.04\% and 9.98\%, and average MAE by
2.42\% and 6.00\%, respectively.
Results on USHCN are mixed: random initialization performs slightly
better at prediction length 1, whereas pre-training yields lower
mean errors at length 12, with only a marginal MAE difference.
Across its two lengths, pre-training reduces MSE by 0.78\% but
increases MAE by 0.18\%.
These results support the benefit of pre-training for adaptation
with limited training data, while showing that its gains depend
on the dataset and prediction length.

\subsection{Limitations}
\label{app:limitations}

Although QiYao-I is trained with a multi-quantile objective, our evaluation focuses on median point forecasts using MSE and MAE, following prior work. These metrics do not establish the calibration of its predictive quantiles. Our pre-training corpus is underexplored. A controlled comparison of pre-training on naturally irregular data, synthetically generated data, and their combination would further clarify their respective contributions to cross-dataset generalization.

\section{More related works}

\paragraph{Time Series Forecasting.}
Time series forecasting has witnessed rapid progress in recent years, spanning benchmarking, multivariate dependency modeling, irregular and missing observations, exogenous information, probabilistic forecasting, and foundation models.
TFB~\cite{qiu2024tfb} establishes a comprehensive and fair benchmark for systematically evaluating forecasting methods across diverse datasets and settings.
To improve multivariate forecasting, DUET~\cite{qiu2025duet} introduces dual clustering to capture temporal and inter-variable structures, CCD~\cite{cheng2026ccd} models cross-correlations through deformable convolutional networks, and SRSNet~\cite{wu2025srsnet} explores selective representation spaces from a patch perspective.
Beyond model architectures, DBLoss~\cite{qiu2025DBLoss} improves forecasting optimization through a decomposition-based objective.
Real-world time series frequently contain missing or irregularly sampled observations, motivating methods specifically designed for such settings.
GinAR+~\cite{11002729} provides an end-to-end framework for multivariate forecasting with missing values, while Merlin~\cite{yu2025merlin} adopts multi-view representation learning to improve robustness under varying missing rates.
For irregular multivariate time series, Liu et al.~\cite{liu2026rethinking} revisit the problem and develop a simple yet effective baseline, ASTGI~\cite{liu2026astgi} captures adaptive spatio-temporal graph interactions among asynchronously observed variables, and Qiu et al.~\cite{qiu2026bridging} jointly model temporal and frequency-domain structures to better characterize irregular observations.
Another important direction incorporates exogenous information into forecasting.
DAG~\cite{qiu2026dag} models dual correlations between target series and exogenous variables, GCGNet~\cite{li2026gcgnet} leverages graph-consistent generative modeling to characterize structured dependencies, and KITE~\cite{cheng2026kite} introduces knowledge-guided probabilistic modeling for forecasting with external variables.
For probabilistic forecasting, K$^2$VAE~\cite{wu2025k2vae} combines Koopman dynamics, Kalman filtering, and variational autoencoding to explicitly characterize predictive uncertainty.
Recent advances in time series foundation models further extend the modeling paradigm toward general-purpose and adaptable temporal representations.
CoRA~\cite{cheng2026cora} introduces a correlation-aware adapter to enhance foundation models for multivariate forecasting, while STAR~\cite{cheng2026star} develops a state-aware adapter for anomaly detection.
Meanwhile, TAB~\cite{qiu2025tab} provides unified benchmarking for time series anomaly detection, and CATCH~\cite{wu2025catch} models channel dependencies through frequency-domain patching.
TimeART~\cite{wu2026timeart} further explores agentic time series reasoning with tool augmentation, indicating a growing trend toward integrating foundation models, external tools, and structured reasoning.
Related sequence modeling studies in other domains have also investigated structured conditioning and controllable generation.
Follow-Your-Pose~\cite{ma2024followpose} and Follow-Your-Emoji~\cite{ma2024followyouremoji} introduce fine-grained control signals for video generation, while LiveLight~\cite{ma2026livelight} studies interactive real-time video relighting; a broader overview of controllable video generation is presented in~\cite{ma2025controllable}.
In industrial time series applications, MetaGNSDformer~\cite{cheng2026metagnsdformer} combines meta-learning, non-stationary modeling, and frequency-aware attention for remaining useful life prediction of lithium-ion batteries.
Collectively, these studies demonstrate a broader transition from conventional forecasting on complete and regularly sampled sequences toward more general time series models capable of handling complex dependencies, irregular observations, auxiliary information, uncertainty, and heterogeneous temporal contexts.

\textbf{Variable Interaction in Foundation Models.}
Cross-variable interaction is important for multivariate time series, but it remains insufficiently explored in time series foundation models. Many models, such as TimesFM2.5~\citep{das2023decoder}, Lag-Llama~\citep{rasul2023lag}, Timer~\citep{liu2024timer}, and Sundial~\citep{liu2025sundial}, AimTS~\citep{chen2025aimts} mainly adopt a channel-independent paradigm. This design limits explicit information exchange among variables. Some models, such as Chronos2~\citep{ansari2025chronos}, Toto2~\citep{khwaja2026toto}, and MOIRAI~\citep{woo2024unified} support multivariate forecasting, but their interactions are usually based on dense attention or aligned time intervals. These designs are more suitable for regular multivariate time series, where variables share the same temporal grid. In irregular multivariate time series, variables are often observed asynchronously and may have different sampling frequencies, and periodic patterns. Dense or fixed interaction may therefore introduce noisy correlations. QiYao-I addresses this problem by constructing dynamic patch-level variable interactions with frequency awareness. Instead of forcing all variables to interact, QiYao-I selects relevant source patches for each target patch, enabling adaptive message passing across asynchronous variables.

\newpage

\section{Full results}

\begin{table*}[h]
\centering
\caption{Per-horizon zero-shot forecasting results on CESNET and APTC. MSE and MAE are reported on the same scales as Table~\ref{tab:zeroshot_results}; lower is better. Best and second-best values are marked in purple and blue, respectively.}
\label{tab:zeroshot_detail_1}
\renewcommand{\arraystretch}{1.12}
\setlength{\tabcolsep}{4pt}
\resizebox{\textwidth}{!}{%
\begin{tabular}{@{}l|cc|cc|cc|cc@{}}
\toprule
\rowcolor{headergray}
\textbf{Method} & \multicolumn{4}{c|}{\textbf{CESNET} $(10^{-3})$} & \multicolumn{4}{c}{\textbf{APTC} $(10^{-5})$} \\
 & \multicolumn{2}{c|}{$P=168$} & \multicolumn{2}{c|}{$P=336$} & \multicolumn{2}{c|}{$P=50$} & \multicolumn{2}{c|}{$P=100$} \\
 & MSE & MAE & MSE & MAE & MSE & MAE & MSE & MAE \\
\midrule
Toto-2 & 22.39 & 92.69 & 22.44 & 92.58 & \second{389.76} & \second{4615.61} & \best{383.29} & \best{4574.92} \\
Chronos-2 & \second{22.08} & 91.23 & \best{21.93} & \second{90.65} & 406.40 & 4708.10 & 397.30 & 4655.40 \\
TimesFM-2.5 & 22.44 & 91.90 & 22.31 & 91.67 & 395.50 & 4640.80 & 387.70 & 4597.10 \\
MOIRAI-2 & 22.18 & \second{90.68} & 22.72 & 91.35 & 405.70 & 4701.10 & 394.90 & 4637.30 \\
Sundial & 22.68 & 92.00 & 22.32 & 92.11 & 414.00 & 4753.70 & 407.10 & 4716.40 \\
TIME-MOE & 22.14 & 93.40 & 23.58 & 97.87 & 406.80 & 4714.20 & 399.40 & 4671.50 \\
Timer & 31.33 & 130.08 & 36.61 & 144.91 & 434.80 & 4961.40 & 578.10 & 5754.10 \\
\midrule
MIRA & 22.49 & 93.28 & 22.24 & 92.28 & 405.20 & 4673.66 & 415.23 & 4727.43 \\
\textbf{QiYao-I} & \best{21.84} & \best{89.99} & \second{22.17} & \best{90.49} & \best{386.72} & \best{4608.78} & \second{383.88} & \second{4585.40} \\
\bottomrule
\end{tabular}%
}
\end{table*}

\begin{table*}[h]
\centering
\caption{Per-horizon zero-shot forecasting results on ClusterTrace and Seabirds. MSE and MAE are reported on the same scales as Table~\ref{tab:zeroshot_results}; lower is better. Best and second-best values are marked in purple and blue, respectively.}
\label{tab:zeroshot_detail_2}
\renewcommand{\arraystretch}{1.12}
\setlength{\tabcolsep}{4pt}
\resizebox{\textwidth}{!}{%
\begin{tabular}{@{}l|cc|cc|cc|cc@{}}
\toprule
\rowcolor{headergray}
\textbf{Method} & \multicolumn{4}{c|}{\textbf{ClusterTrace} $(10^{-3})$} & \multicolumn{4}{c}{\textbf{Seabirds} $(10^{-6})$} \\
 & \multicolumn{2}{c|}{$P=24$} & \multicolumn{2}{c|}{$P=48$} & \multicolumn{2}{c|}{$P=360$} & \multicolumn{2}{c|}{$P=720$} \\
 & MSE & MAE & MSE & MAE & MSE & MAE & MSE & MAE \\
\midrule
Toto-2 & \second{38.70} & \second{141.20} & \second{44.46} & \second{156.83} & 184.55 & \second{6618.25} & 442.25 & 11443.38 \\
Chronos-2 & 42.17 & 149.08 & 49.81 & 165.78 & \second{175.00} & 6940.00 & \second{416.00} & 11828.00 \\
TimesFM-2.5 & 41.98 & 149.71 & 50.68 & 169.95 & 233.00 & 7565.00 & 483.00 & 11916.00 \\
MOIRAI-2 & 45.61 & 163.02 & 62.12 & 192.63 & 195.00 & 7061.00 & 417.00 & \second{10730.00} \\
Sundial & 46.62 & 159.84 & 48.79 & 166.31 & 249.00 & 9435.00 & 433.00 & 12950.00 \\
TIME-MOE & 51.91 & 167.87 & 49.72 & 165.72 & 416.00 & 13015.00 & 628.00 & 17067.00 \\
Timer & 118.33 & 240.93 & 174.75 & 320.17 & 26911.00 & 91971.00 & 83126.00 & 172879.00 \\
\midrule
MIRA & 45.75 & 159.39 & 51.96 & 174.51 & 396.57 & 13308.31 & 542.64 & 16407.14 \\
\textbf{QiYao-I} & \best{31.97} & \best{132.94} & \best{42.46} & \best{152.41} & \best{124.58} & \best{6450.75} & \best{206.77} & \best{8372.78} \\
\bottomrule
\end{tabular}%
}
\end{table*}

\begin{table*}[h]
\centering
\caption{Per-horizon zero-shot forecasting results on Human Activity and FNSPID. MSE and MAE are reported on the same scales as Table~\ref{tab:zeroshot_results}; lower is better. Best and second-best values are marked in purple and blue, respectively.}
\label{tab:zeroshot_detail_3}
\renewcommand{\arraystretch}{1.12}
\setlength{\tabcolsep}{4pt}
\resizebox{\textwidth}{!}{%
\begin{tabular}{@{}l|cc|cc|cc|cc@{}}
\toprule
\rowcolor{headergray}
\textbf{Method} & \multicolumn{4}{c|}{\textbf{Human Activity} $(10^{-4})$} & \multicolumn{4}{c}{\textbf{FNSPID} $(10^{-5})$} \\
 & \multicolumn{2}{c|}{$P=200$} & \multicolumn{2}{c|}{$P=3000$} & \multicolumn{2}{c|}{$P=180$} & \multicolumn{2}{c|}{$P=365$} \\
 & MSE & MAE & MSE & MAE & MSE & MAE & MSE & MAE \\
\midrule
Toto-2 & 31.82 & \best{301.85} & 54.08 & \best{456.17} & \second{166.14} & \second{2365.01} & \second{395.63} & \second{3689.53} \\
Chronos-2 & 31.14 & 306.45 & \second{53.72} & \second{466.47} & 191.20 & 2543.40 & 430.80 & 3963.80 \\
TimesFM-2.5 & 39.80 & 334.90 & 71.47 & 534.36 & 220.90 & 2826.60 & 543.60 & 4376.10 \\
MOIRAI-2 & 32.94 & 309.99 & 63.17 & 496.02 & 201.60 & 2653.10 & 561.30 & 4531.90 \\
Sundial & \second{30.21} & \second{303.81} & 58.92 & 497.23 & 257.10 & 3111.80 & 585.60 & 4663.60 \\
TIME-MOE & 32.78 & 331.21 & 82.44 & 605.23 & 236.80 & 2990.30 & 534.60 & 4418.20 \\
Timer & 54.90 & 450.58 & 650.40 & 1800.49 & 7772.30 & 18312.50 & 20111.60 & 29413.20 \\
\midrule
MIRA & 707.47 & 2222.64 & 1127.21 & 3112.64 & 465.51 & 4568.12 & 429.98 & 4488.40 \\
\textbf{QiYao-I} & \best{29.20} & 309.64 & \best{53.71} & 474.56 & \best{148.23} & \best{2358.64} & \best{351.37} & \best{3673.60} \\
\bottomrule
\end{tabular}%
}
\end{table*}

\begin{table*}[h]
\centering
\caption{Per-horizon zero-shot forecasting results on GDELT and USHCN. MSE and MAE are reported on the same scales as Table~\ref{tab:zeroshot_results}; lower is better. Best and second-best values are marked in purple and blue, respectively. USHCN uses z-score normalization; its ranking includes only models with verified per-horizon results.}
\label{tab:zeroshot_detail_4}
\renewcommand{\arraystretch}{1.12}
\setlength{\tabcolsep}{4pt}
\resizebox{\textwidth}{!}{%
\begin{tabular}{@{}l|cc|cc|cc|cc@{}}
\toprule
\rowcolor{headergray}
\textbf{Method} & \multicolumn{4}{c|}{\textbf{GDELT} $(10^{-4})$} & \multicolumn{4}{c}{\textbf{USHCN} $(10^{-3})$} \\
 & \multicolumn{2}{c|}{$P=30$} & \multicolumn{2}{c|}{$P=90$} & \multicolumn{2}{c|}{$P=1$} & \multicolumn{2}{c|}{$P=12$} \\
 & MSE & MAE & MSE & MAE & MSE & MAE & MSE & MAE \\
\midrule
Toto-2 & 201.56 & \second{847.14} & 196.25 & \second{836.48} & 1166.67 & 432.62 & 961.55 & 443.17 \\
Chronos-2 & 202.21 & 851.26 & \second{194.66} & \best{834.86} & 1095.07 & 419.62 & 1151.71 & \best{436.62} \\
TimesFM-2.5 & 205.71 & 853.00 & 203.03 & 842.42 & 1302.10 & 461.30 & 1712.54 & 587.61 \\
MOIRAI-2 & 204.85 & 861.93 & 203.00 & 857.98 & 1130.38 & 440.32 & 1275.23 & 529.75 \\
Sundial & 209.39 & 881.07 & 198.71 & 853.37 & 1042.24 & \second{415.81} & \second{895.61} & \second{442.50} \\
TIME-MOE & \second{198.94} & 869.98 & \best{194.07} & 865.08 & \best{859.57} & \best{403.62} & 1126.11 & 533.82 \\
Timer & 232.91 & 1030.10 & 252.94 & 1089.89 & 1442.00 & 558.05 & 1231.00 & 493.00 \\
\midrule
MIRA & 402.70 & 1328.72 & 432.10 & 1315.03 & 1052.04 & 485.50 & 1225.61 & 502.11 \\
\textbf{QiYao-I} & \best{196.71} & \best{844.94} & 198.23 & 838.37 & \second{928.80} & 459.12 & \best{886.73} & 454.97 \\
\bottomrule
\end{tabular}%
}
\end{table*}


\begin{table*}[h]
\centering
\caption{Per-horizon forecasting results on CESNET and APTC. QiYao-I uses 10\% few-shot training; the end-to-end baselines use full-shot training. MSE and MAE use the scales of the main few-shot table. The best and second-best reported values are highlighted in purple and blue, respectively; ties at the reported precision are highlighted equally.}
\label{tab:fewshot_detail_1}
\renewcommand{\arraystretch}{1.10}
\setlength{\tabcolsep}{4pt}
\resizebox{\textwidth}{!}{%
\begin{tabular}{@{}l|cc|cc|cc|cc@{}}
\toprule
\rowcolor{headergray}
\textbf{Method} & \multicolumn{4}{c|}{\textbf{CESNET} $(10^{-3})$} & \multicolumn{4}{c}{\textbf{APTC} $(10^{-5})$} \\
 & \multicolumn{2}{c|}{$P=168$} & \multicolumn{2}{c|}{$P=336$} & \multicolumn{2}{c|}{$P=50$} & \multicolumn{2}{c|}{$P=100$} \\
 & MSE & MAE & MSE & MAE & MSE & MAE & MSE & MAE \\
\midrule
TFMixer & 23.30 & 100.30 & 23.60 & 102.60 & \second{390.00} & \second{4680.00} & \second{380.00} & \second{4580.00} \\
ASTGI & \second{21.60} & \second{92.10} & \best{21.60} & \second{94.30} & \second{390.00} & \second{4680.00} & \second{380.00} & 4630.00 \\
APN & 23.10 & 101.70 & 23.50 & 101.00 & \second{390.00} & 4690.00 & \second{380.00} & 4590.00 \\
KAFNet & 23.60 & 103.70 & 23.50 & 104.10 & \second{390.00} & \second{4680.00} & \second{380.00} & \second{4580.00} \\
Hi-Patch & 24.40 & 103.70 & 25.20 & 104.40 & \second{390.00} & \second{4680.00} & \second{380.00} & \second{4580.00} \\
HyperIMTS & 27.00 & 122.80 & 29.80 & 133.40 & \second{390.00} & \second{4680.00} & \second{380.00} & \second{4580.00} \\
GNeuralFlow & 23.40 & 103.90 & 23.50 & 103.60 & \second{390.00} & 4710.00 & \second{380.00} & 4590.00 \\
tPatchGNN & 23.40 & 102.70 & 23.50 & 103.50 & 560.00 & 5740.00 & 620.00 & 6080.00 \\
GraFITi & 23.10 & 104.20 & 23.00 & 103.90 & \second{390.00} & \second{4680.00} & \second{380.00} & \second{4580.00} \\
Warpformer & 23.40 & 101.60 & 41.00 & 136.00 & 570.00 & 5790.00 & 630.00 & 6150.00 \\
CRU & 23.30 & 103.80 & 23.80 & 106.90 & \second{390.00} & \second{4680.00} & \second{380.00} & \second{4580.00} \\
NeuralFlow & 23.50 & 104.00 & 23.40 & 102.90 & \second{390.00} & 4700.00 & \second{380.00} & 4600.00 \\
mTAN & 23.50 & 103.80 & 23.20 & 103.70 & \second{390.00} & \second{4680.00} & \second{380.00} & \second{4580.00} \\
SeFT & 50.80 & 154.00 & 50.80 & 154.40 & 430.00 & 4970.00 & 440.00 & 5080.00 \\
GRU-D & 23.30 & 103.10 & 23.30 & 103.30 & \second{390.00} & \second{4680.00} & \second{380.00} & \second{4580.00} \\
\midrule
\textbf{QiYao-I (10\%)} & \best{21.34} & \best{87.08} & \second{21.83} & \best{87.90} & \best{369.84} & \best{4504.04} & \best{371.99} & \best{4510.99} \\
\bottomrule
\end{tabular}%
}
\end{table*}

\begin{table*}[h]
\centering
\caption{Per-horizon forecasting results on ClusterTrace and Seabirds. QiYao-I uses 10\% few-shot training; the end-to-end baselines use full-shot training. MSE and MAE use the scales of the main few-shot table. The best and second-best reported values are highlighted in purple and blue, respectively; ties at the reported precision are highlighted equally.}
\label{tab:fewshot_detail_2}
\renewcommand{\arraystretch}{1.10}
\setlength{\tabcolsep}{4pt}
\resizebox{\textwidth}{!}{%
\begin{tabular}{@{}l|cc|cc|cc|cc@{}}
\toprule
\rowcolor{headergray}
\textbf{Method} & \multicolumn{4}{c|}{\textbf{ClusterTrace} $(10^{-3})$} & \multicolumn{4}{c}{\textbf{Seabirds} $(10^{-6})$} \\
 & \multicolumn{2}{c|}{$P=24$} & \multicolumn{2}{c|}{$P=48$} & \multicolumn{2}{c|}{$P=360$} & \multicolumn{2}{c|}{$P=720$} \\
 & MSE & MAE & MSE & MAE & MSE & MAE & MSE & MAE \\
\midrule
TFMixer & \second{39.20} & \second{147.30} & 43.60 & 153.00 & 300.00 & 13400.00 & 500.00 & 15800.00 \\
ASTGI & 40.60 & 157.50 & \second{42.50} & \best{152.30} & \second{200.00} & \second{11700.00} & \second{400.00} & \second{13300.00} \\
APN & 42.50 & 153.10 & 44.70 & 154.10 & 700.00 & 19400.00 & 1400.00 & 32800.00 \\
KAFNet & 45.00 & 155.00 & 44.90 & 154.80 & 900.00 & 28300.00 & 1100.00 & 27300.00 \\
Hi-Patch & 43.70 & 156.20 & 42.80 & 153.40 & 24800.00 & 132900.00 & 193000.00 & 430100.00 \\
HyperIMTS & 44.70 & 157.50 & 43.20 & 158.60 & 500.00 & 16300.00 & \second{400.00} & 13800.00 \\
GNeuralFlow & 47.20 & 163.20 & 44.70 & 156.70 & 13700.00 & 96000.00 & 16200.00 & 105500.00 \\
tPatchGNN & 46.20 & 157.30 & 44.60 & 154.10 & 1000.00 & 22500.00 & 1400.00 & 26200.00 \\
GraFITi & 44.80 & 155.90 & 44.60 & 154.70 & 500.00 & 18000.00 & 700.00 & 22200.00 \\
Warpformer & 86.00 & 220.00 & 97.10 & 238.70 & 32300.00 & 151900.00 & 251900.00 & 442400.00 \\
CRU & 48.50 & 171.70 & 48.00 & 160.10 & 6300.00 & 55200.00 & 5800.00 & 53800.00 \\
NeuralFlow & 44.70 & 155.50 & 44.70 & 155.60 & 1700.00 & 30200.00 & 1400.00 & 29700.00 \\
mTAN & 50.60 & 176.20 & 49.20 & 161.20 & 4500.00 & 42900.00 & 700.00 & 20000.00 \\
SeFT & 99.50 & 244.60 & 99.80 & 244.80 & 8400.00 & 88500.00 & 8600.00 & 83500.00 \\
GRU-D & 46.30 & 157.20 & 44.60 & 154.20 & 8800.00 & 59600.00 & 47500.00 & 180300.00 \\
\midrule
\textbf{QiYao-I (10\%)} & \best{32.67} & \best{132.90} & \best{40.48} & \second{152.71} & \best{162.70} & \best{7674.32} & \best{180.09} & \best{7946.82} \\
\bottomrule
\end{tabular}%
}
\end{table*}

\begin{table*}[h]
\centering
\caption{Per-horizon forecasting results on Human Activity and FNSPID. QiYao-I uses 10\% few-shot training; the end-to-end baselines use full-shot training. MSE and MAE use the scales of the main few-shot table. The best and second-best reported values are highlighted in purple and blue, respectively; ties at the reported precision are highlighted equally.}
\label{tab:fewshot_detail_3}
\renewcommand{\arraystretch}{1.10}
\setlength{\tabcolsep}{4pt}
\resizebox{\textwidth}{!}{%
\begin{tabular}{@{}l|cc|cc|cc|cc@{}}
\toprule
\rowcolor{headergray}
\textbf{Method} & \multicolumn{4}{c|}{\textbf{Human Activity} $(10^{-4})$} & \multicolumn{4}{c}{\textbf{FNSPID} $(10^{-5})$} \\
 & \multicolumn{2}{c|}{$P=200$} & \multicolumn{2}{c|}{$P=3000$} & \multicolumn{2}{c|}{$P=180$} & \multicolumn{2}{c|}{$P=365$} \\
 & MSE & MAE & MSE & MAE & MSE & MAE & MSE & MAE \\
\midrule
TFMixer & \best{25.00} & \best{281.00} & \best{48.00} & \best{443.00} & \best{200.00} & \second{3210.00} & 640.00 & 5470.00 \\
ASTGI & 28.00 & 317.00 & 51.00 & 484.00 & 720.00 & 7240.00 & 1000.00 & 8450.00 \\
APN & 26.00 & 297.00 & \best{48.00} & 453.00 & 320.00 & 4080.00 & 2430.00 & 12210.00 \\
KAFNet & 27.00 & 305.00 & 51.00 & 483.00 & 350.00 & 4430.00 & \best{400.00} & \best{4900.00} \\
Hi-Patch & 29.00 & 337.00 & \second{49.00} & 486.00 & 660.00 & 5490.00 & 3930.00 & 15320.00 \\
HyperIMTS & 26.00 & 303.00 & 54.00 & 515.00 & 560.00 & 5470.00 & \second{530.00} & 5300.00 \\
GNeuralFlow & 185.00 & 1064.00 & 179.00 & 1041.00 & 2910.00 & 13690.00 & 1070.00 & 7590.00 \\
tPatchGNN & 27.00 & 303.00 & \best{48.00} & 472.00 & 640.00 & 5600.00 & 1190.00 & 9240.00 \\
GraFITi & 26.00 & 296.00 & 50.00 & 479.00 & 660.00 & 5990.00 & 1460.00 & 9170.00 \\
Warpformer & 40.00 & 436.00 & 208.00 & 1132.00 & 3770.00 & 16140.00 & 3690.00 & 15590.00 \\
CRU & 82.00 & 645.00 & 100.00 & 734.00 & 1010.00 & 7980.00 & 1070.00 & 7890.00 \\
NeuralFlow & 98.00 & 752.00 & 102.00 & 772.00 & 760.00 & 6520.00 & 610.00 & 5780.00 \\
mTAN & 52.00 & 519.00 & 70.00 & 619.00 & 890.00 & 7130.00 & 800.00 & 7320.00 \\
SeFT & 164.00 & 1019.00 & 168.00 & 1025.00 & 3690.00 & 16480.00 & 3650.00 & 15830.00 \\
GRU-D & 40.00 & 434.00 & 57.00 & 550.00 & 1290.00 & 9480.00 & 1880.00 & 10440.00 \\
\midrule
\textbf{QiYao-I (10\%)} & \second{25.75} & \second{287.07} & 53.28 & \second{450.51} & \second{258.26} & \best{2905.11} & 694.00 & \second{4927.13} \\
\bottomrule
\end{tabular}%
}
\end{table*}

\begin{table*}[h]
\centering
\caption{Per-horizon forecasting results on GDELT and USHCN. QiYao-I uses 10\% few-shot training; the end-to-end baselines use full-shot training. MSE and MAE use the scales of the main few-shot table. The best and second-best reported values are highlighted in purple and blue, respectively; ties at the reported precision are highlighted equally. USHCN uses z-score normalization.}
\label{tab:fewshot_detail_4}
\renewcommand{\arraystretch}{1.10}
\setlength{\tabcolsep}{4pt}
\resizebox{\textwidth}{!}{%
\begin{tabular}{@{}l|cc|cc|cc|cc@{}}
\toprule
\rowcolor{headergray}
\textbf{Method} & \multicolumn{4}{c|}{\textbf{GDELT} $(10^{-4})$} & \multicolumn{4}{c}{\textbf{USHCN} $(10^{-3})$} \\
 & \multicolumn{2}{c|}{$P=30$} & \multicolumn{2}{c|}{$P=90$} & \multicolumn{2}{c|}{$P=1$} & \multicolumn{2}{c|}{$P=12$} \\
 & MSE & MAE & MSE & MAE & MSE & MAE & MSE & MAE \\
\midrule
TFMixer & 191.00 & \second{843.00} & 192.00 & 848.00 & \second{537.60} & \best{251.50} & \best{491.80} & \second{277.90} \\
ASTGI & \best{188.00} & \best{840.00} & \second{188.00} & \second{841.00} & 582.80 & 370.90 & \second{497.80} & 335.80 \\
APN & 190.00 & 851.00 & 190.00 & 860.00 & \best{523.00} & 309.70 & 514.20 & 357.00 \\
KAFNet & 189.00 & 859.00 & 189.00 & 861.00 & 583.80 & 351.60 & 697.30 & 478.10 \\
Hi-Patch & 195.00 & 886.00 & 190.00 & 873.00 & 571.90 & 369.70 & 541.60 & 403.70 \\
HyperIMTS & \best{188.00} & 848.00 & 191.00 & 851.00 & 662.40 & 505.90 & 721.70 & 466.80 \\
GNeuralFlow & 190.00 & 858.00 & 191.00 & 862.00 & 547.30 & 354.90 & 710.30 & 466.20 \\
tPatchGNN & 191.00 & \best{840.00} & 192.00 & \best{831.00} & 577.60 & 363.90 & 753.60 & 510.90 \\
GraFITi & \best{188.00} & 848.00 & 211.00 & 990.00 & 559.30 & 333.40 & 718.30 & 522.90 \\
Warpformer & 1610.00 & 3332.00 & 1469.00 & 3174.00 & 552.50 & 409.00 & 537.40 & 372.80 \\
CRU & 190.00 & 844.00 & 196.00 & 883.00 & 674.10 & 428.10 & 617.70 & 409.40 \\
NeuralFlow & 190.00 & 848.00 & 190.00 & 858.00 & 564.30 & 350.60 & 578.60 & 393.70 \\
mTAN & 205.00 & 897.00 & 192.00 & 858.00 & 1069.30 & 651.50 & 587.00 & 366.70 \\
SeFT & 1388.00 & 2989.00 & 1765.00 & 3145.00 & 799.50 & 503.70 & 751.20 & 505.40 \\
GRU-D & 189.00 & 849.00 & 189.00 & 854.00 & 578.70 & 365.20 & 750.10 & 487.70 \\
\midrule
\textbf{QiYao-I (10\%)} & \second{188.23} & 848.59 & \best{187.93} & 848.03 & 560.59 & \second{260.23} & 517.05 & \best{246.45} \\
\bottomrule
\end{tabular}%
}
\end{table*}

\end{document}